\documentclass[sn-basic,iicol,Numbered]{sn-jnl}

\usepackage{graphicx}%
\usepackage{multirow}%
\usepackage{amsmath,amssymb,amsfonts}%
\usepackage{amsthm}%
\usepackage{mathrsfs}%
\usepackage[title]{appendix}%
\usepackage{xcolor}%
\usepackage{textcomp}%
\usepackage{manyfoot}%
\usepackage{booktabs}%
\usepackage{algorithm}%
\usepackage{algorithmicx}%
\usepackage{algpseudocode}%
\usepackage{listings}%

\usepackage{paralist}%
\usepackage[
singlelinecheck=false 
]{caption}
\usepackage{stfloats}
\usepackage{pifont}
\usepackage{multirow}
\usepackage{subfig}

\theoremstyle{thmstyleone}%
\theoremstyle{thmstyletwo}%

\theoremstyle{thmstylethree}%

\begin{document}

\title{FabuLight-ASD: Unveiling Speech Activity via Body Language}

\author*{\fnm{Hugo} \sur{Carneiro}}\email{hugo.carneiro@uni-hamburg.de}

\author{\fnm{Stefan} \sur{Wermter}}\email{stefan.wermter@uni-hamburg.de}

\affil{\orgdiv{Department of Informatics, Knowledge Technology}, \orgname{Universit{\"a}t Hamburg}, \orgaddress{\street{Vogt-Koelln-Stra{\ss}e 30}, \city{Hamburg}, \postcode{22527}, \country{Germany}}}

\abstract{Active speaker detection (ASD) in multimodal environments is crucial for various applications, from video conferencing to human-robot interaction. This paper introduces FabuLight-ASD, an advanced ASD model that integrates facial, audio, and body pose information to enhance detection accuracy and robustness. Our model builds upon the existing Light-ASD framework by incorporating human pose data, represented through skeleton graphs, which minimises computational overhead. Using the Wilder Active Speaker Detection (WASD) dataset, renowned for reliable face and body bounding box annotations, we demonstrate FabuLight-ASD's effectiveness in real-world scenarios. Achieving an overall mean average precision (mAP) of 94.3\%, FabuLight-ASD outperforms Light-ASD, which has an overall mAP of 93.7\% across various challenging scenarios. The incorporation of body pose information shows a particularly advantageous impact, with notable improvements in mAP observed in scenarios with speech impairment, face occlusion, and human voice background noise. Furthermore, efficiency analysis indicates only a modest increase in parameter count (27.3\%) and multiply-accumulate operations (up to 2.4\%), underscoring the model's efficiency and feasibility. These findings validate the efficacy of FabuLight-ASD in enhancing ASD performance through the integration of body pose data. FabuLight-ASD's code and model weights are available at \url{https://github.com/knowledgetechnologyuhh/FabuLight-ASD}.}




\maketitle

\section{Introduction}\label{sec:intro}

    Active speaker detection (ASD) aims to determine whether a specific person within a video scene is speaking or silent in each frame. This task is essential for various applications, such as speaker diarisation~\cite{chung2019a,chung2020}, speech enhancement~\cite{afouras2018}, speaker localisation and tracking~\cite{carneiro2023,qian2021,qian2022}, speech separation~\cite{qu2020}, and human-robot interaction~\cite{stefanov2016,stefanov2017}. By accurately identifying when a person is speaking or silent within a given timeframe, ASD enables the extraction of valuable insights from audiovisual data, driving advancements across multiple domains.

\begin{figure*}[!ht]
        \centering
        \captionsetup{justification=justified}
        \includegraphics[width=0.925\textwidth]{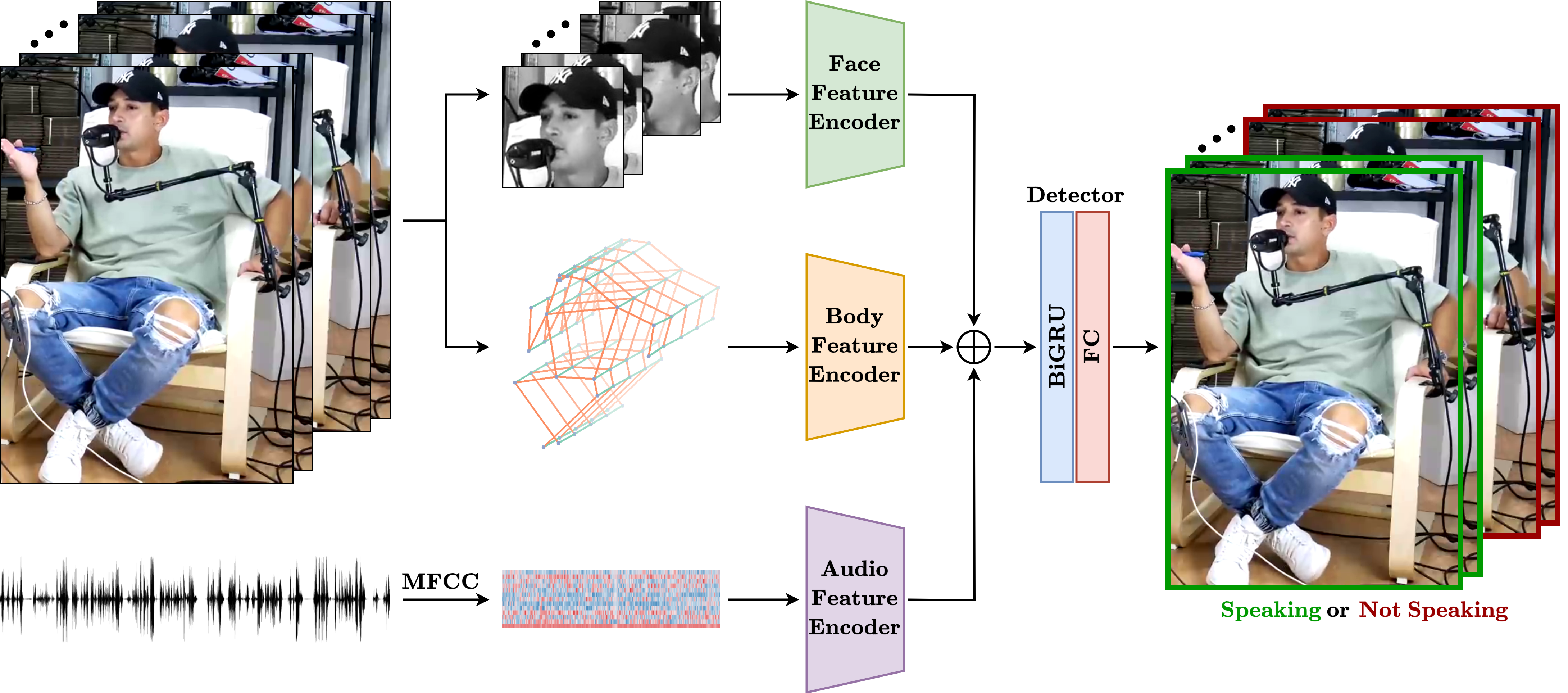}\vspace{0.15cm}
        \caption{The architecture of FabuLight-ASD. The model determines whether a target individual is actively speaking or silent in each video frame based on the face crops and body poses of that individual across all frames, along with the corresponding audio information. The overall architecture builds on Light-ASD, with the body feature encoder being a newly added component that improves performance.}
        \label{fig:fabulightasd-architecture}
    \end{figure*}

    Recent advancements in lightweight and efficient ASD approaches, such as Light-ASD~\cite{liao2023}, have shown promise for deployment in embedded architectures, such as social robots. These approaches allow fast and accurate determination of active speakers within a group, facilitating more fluid and credible human-robot interactions. However, in scenarios involving large groups or where individuals are at a distance from the robot, the effectiveness of ASD models that rely solely on facial and audio cues is limited. To overcome this challenge, we propose FabuLight-ASD (\textbf{F}ace, \textbf{a}udio, and \textbf{b}ody \textbf{u}tilisation for \textbf{Light}weight \textbf{A}ctive \textbf{S}peaker \textbf{D}etection), an extension of Light-ASD that integrates skeleton-based pose information from the target individual. Figure~\ref{fig:fabulightasd-architecture} presents FabuLight-ASD's architecture. This enhancement allows the model to complement cues from facial expressions and audio with body pose information, improving its ability to accurately identify active speakers, even in scenarios where facial nuances are not easily discernible. By leveraging multiple modalities, FabuLight-ASD aims to improve the robustness and accuracy of speaker detection, thereby advancing the capabilities of ASD systems for deployment in real-world scenarios, including those involving social robots.
    
\begin{figure*}[!ht]
        \centering
        \captionsetup{justification=centering}
        \subfloat[Optimal conditions]{\includegraphics[height=0.13625\textheight]{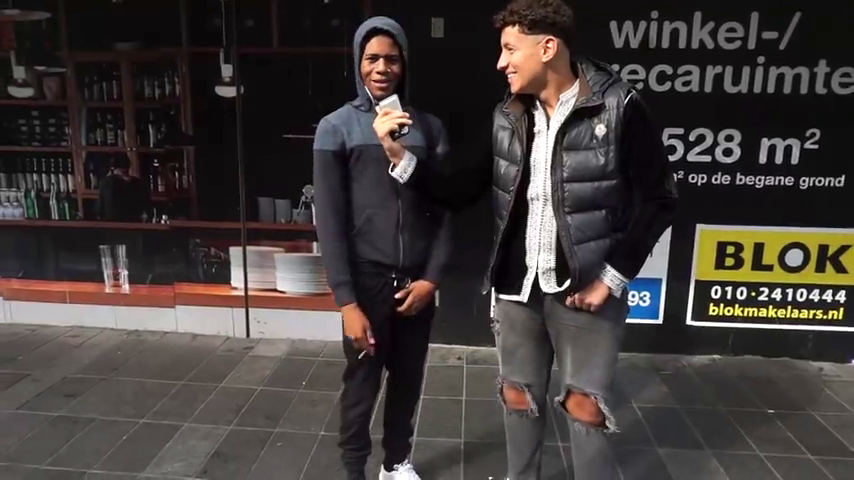}\label{fig:asd-wasd_oc}}\hspace*{0.15cm}
        \subfloat[Speech impairment]{\includegraphics[height=0.13625\textheight]{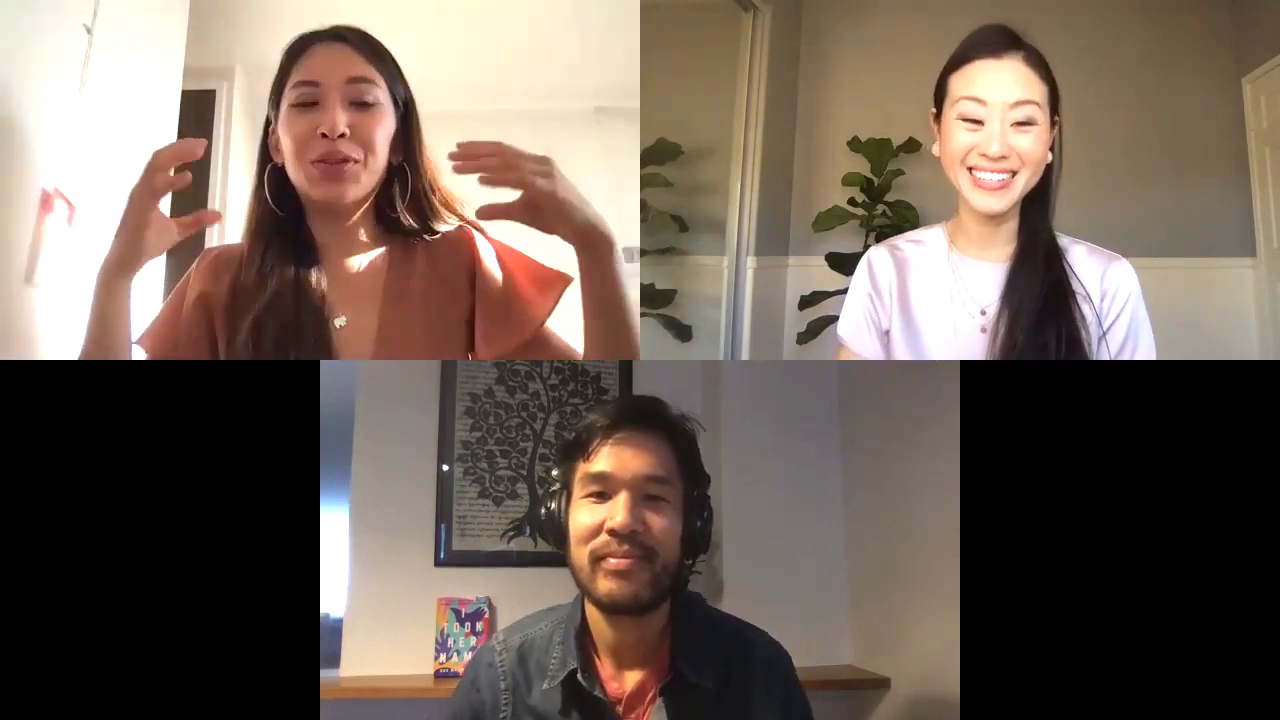}\label{fig:asd-wasd_si}}\hspace*{0.15cm}
        \subfloat[Face occlusion]{\includegraphics[height=0.13625\textheight]{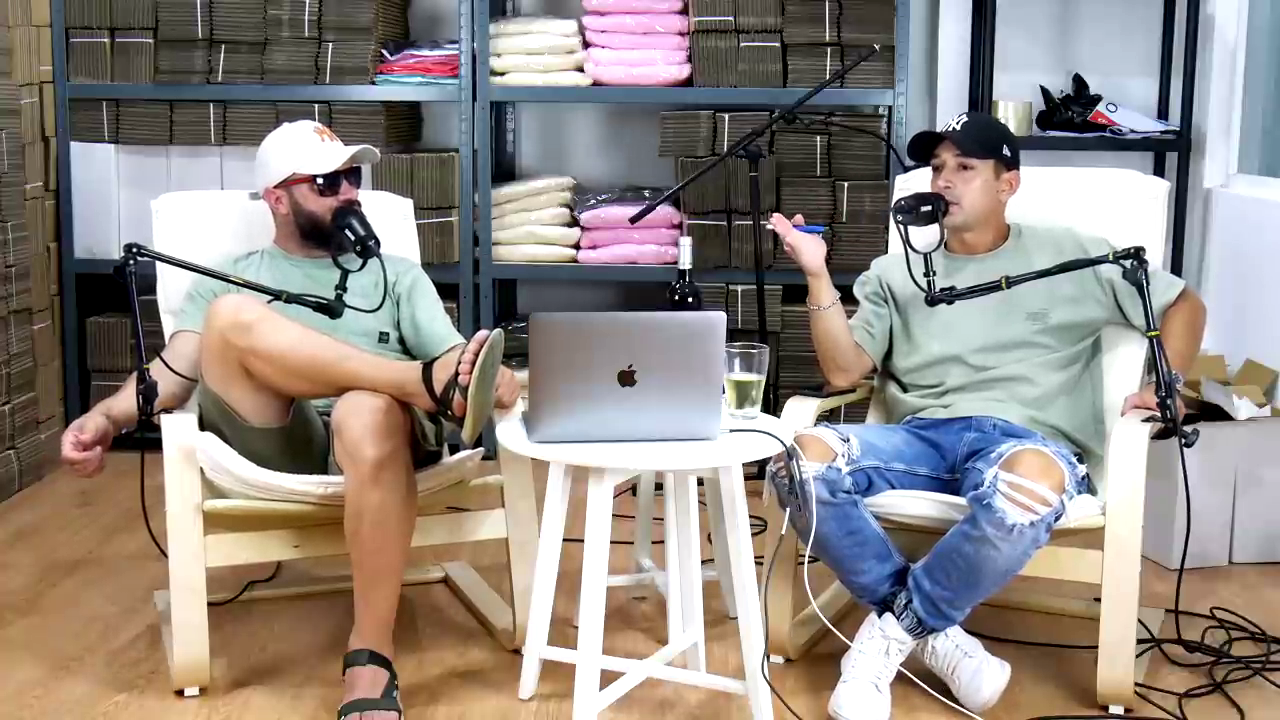}\label{fig:asd-wasd_fo}}\\
        \subfloat[Human voice noise]{\includegraphics[height=0.13625\textheight]{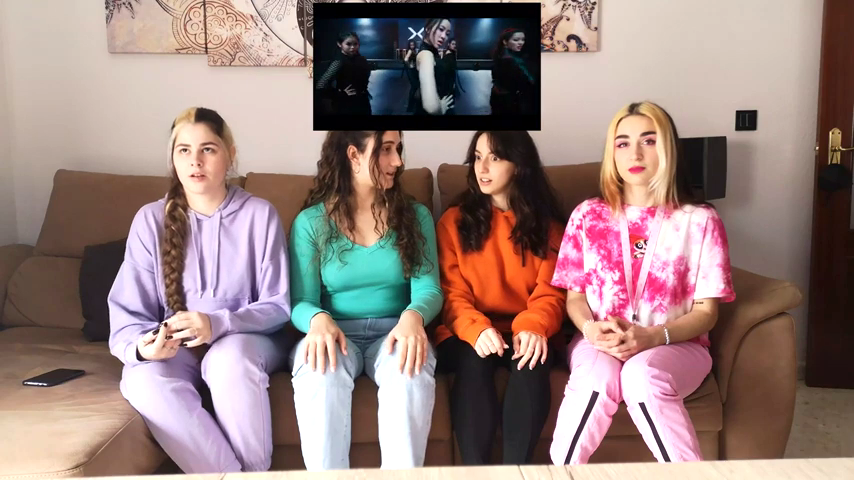}\label{fig:asd-wasd_hvn}}\hspace*{0.15cm}
        \subfloat[Surveillance settings]{\includegraphics[height=0.13625\textheight]{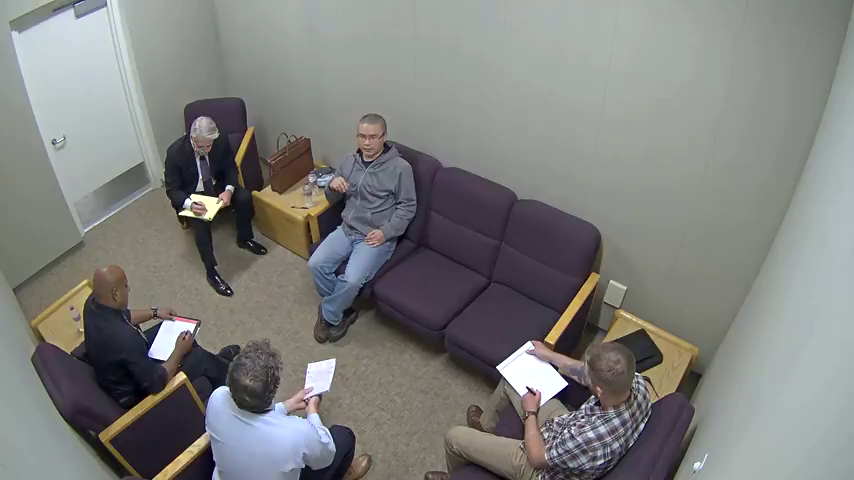}\label{fig:asd-wasd_ss}}\vspace{0.15cm}
        \caption{Examples of WASD videos of various categories.}
        \label{fig:wasd_categories}
    \end{figure*}

    To capitalise on pose information for active speaker detection, we opted to utilise the WASD (Wilder Active Speaker Detection) dataset~\cite{roxo2024} as a benchmark. WASD stands out for its reliable face position annotations and comprehensive body position annotations. Moreover, WASD presents a diverse range of challenging scenarios, making it conducive to the development of more robust ASD models. Notably, WASD features a high frequency of speaking instances, which is beneficial for applications requiring ASD models that are not biased toward non-speaking individuals~\cite{carneiro2023}. Figure~\ref{fig:wasd_categories} provides examples of the various challenging scenarios in WASD. 

    The contributions of this paper are threefold:
    \begin{inparaenum}[\bfseries(1)\upshape]
        \item development of FabuLight-ASD, an ASD model that integrates facial, audio, and pose information to enhance speaker detection performance and robustness;
        \item demonstration of the relevance of human pose to detect the source of speaking activity in challenging scenarios; and
        \item quantification of FabuLight-ASD's efficiency in terms of model size and computational load.
    \end{inparaenum}
    
    The structure of the paper is outlined as follows: Section~\ref{sec:asd} provides an overview of the datasets and existing approaches to the active speaker detection task, highlighting their limitations and discussing their potential for extension to incorporate skeleton-based pose information as an additional input modality. In Section~\ref{sec:lightasd}, we examine in detail the architecture of Light-ASD. In Section~\ref{sec:fabulightasd}, we introduce FabuLight-ASD, our proposed approach, detailing how it integrates human body pose information into the architecture inherited from Light-ASD, and the necessary architectural adjustments for this integration. Section~\ref{sec:experiments} presents the experiments and subsequent analyses conducted to assess the impact of pose information on ASD performance. Section~\ref{sec:conclusion} summarises the paper's contributions and suggests directions for future research.
    
\section{Active Speaker Detection}\label{sec:asd}
    
    The task of active speaker detection (ASD) traces back to the pioneering work of Cutler and Davis~\cite{cutler2000}, who employ a time-delayed network to learn audiovisual correlations from speech activity. Historically, ASD solutions relied on small, handcrafted, and task-specific datasets. To address the lack of an in-the-wild dataset for ASD, Roth et al.~\cite{roth2020} introduce AVA-ActiveSpeaker (Atomic Visual Action -- ActiveSpeaker), the first large-scale, task-agnostic dataset for active speaker detection. This dataset comprises video footage in various languages and resolutions, depicting individual faces from different angles. It was initially released for the ActivityNet Challenge 2019\footnote{\url{http://activity-net.org/challenges/2019/tasks/guest_ava.html}}, where competing models were ranked based on their mean average precision (mAP). Since then, mAP has become the standard metric for comparing ASD models. Each record in AVA-ActiveSpeaker representing an individual in a video frame, assigned a label indicating whether the person is
    \begin{inparaenum}[(i)]
        \item not speaking, 
        \item speaking with their voice audible, or
        \item speaking with other audio overlaying their voice.
    \end{inparaenum}

    Since the release of AVA-ActiveSpeaker, numerous approaches have been proposed to tackle in-the-wild ASD~\cite{alcazar2020,alcazar2021,alcazar2022,carneiro2021,chung2019b,datta2022,jiang2023,kopuklu2021,liao2023,min2022,radman2024,shahid2021,tao2021,werkaixi2022,xiong2023,zhang2019,zhang2021a,zhang2021b,zhang2022}. However, all of these models have shown significant performance decrease when dealing with small face crops (face width smaller than 64 pixels). In scenarios where face information is unreliable, such as low resolution, occlusion, or when containing non-talking lip movements, body pose information can provide additional cues to disambiguate these cases~\cite{roxo2024}, owing to the correlation between upper-body limb movements and speech activity~\cite{hedge2023}.
    
    \begin{figure*}[!ht]
        \centering
        \captionsetup{justification=centering}
        \subfloat[Not a person]{\includegraphics[height=0.15\textheight]{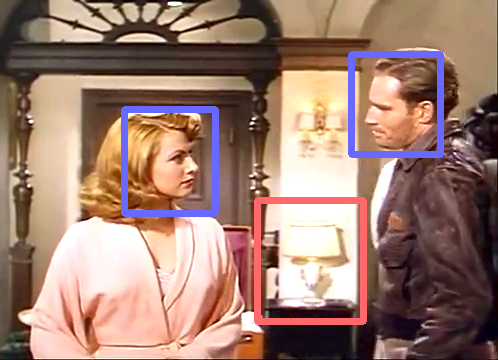}\label{fig:ava_not_a_person}}\hspace*{3cm}
        \subfloat[Misplaced bounding box]{\includegraphics[height=0.15\textheight]{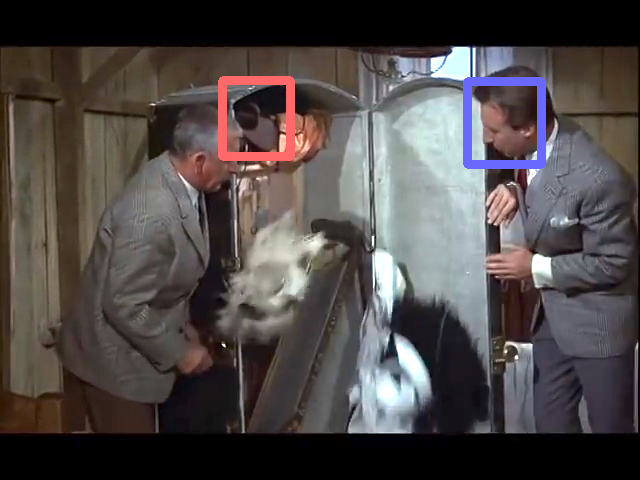}\label{fig:ava_misplaced_bbox}}\\
        \subfloat[Incorrect normalised coordinates]{\includegraphics[height=0.15\textheight]{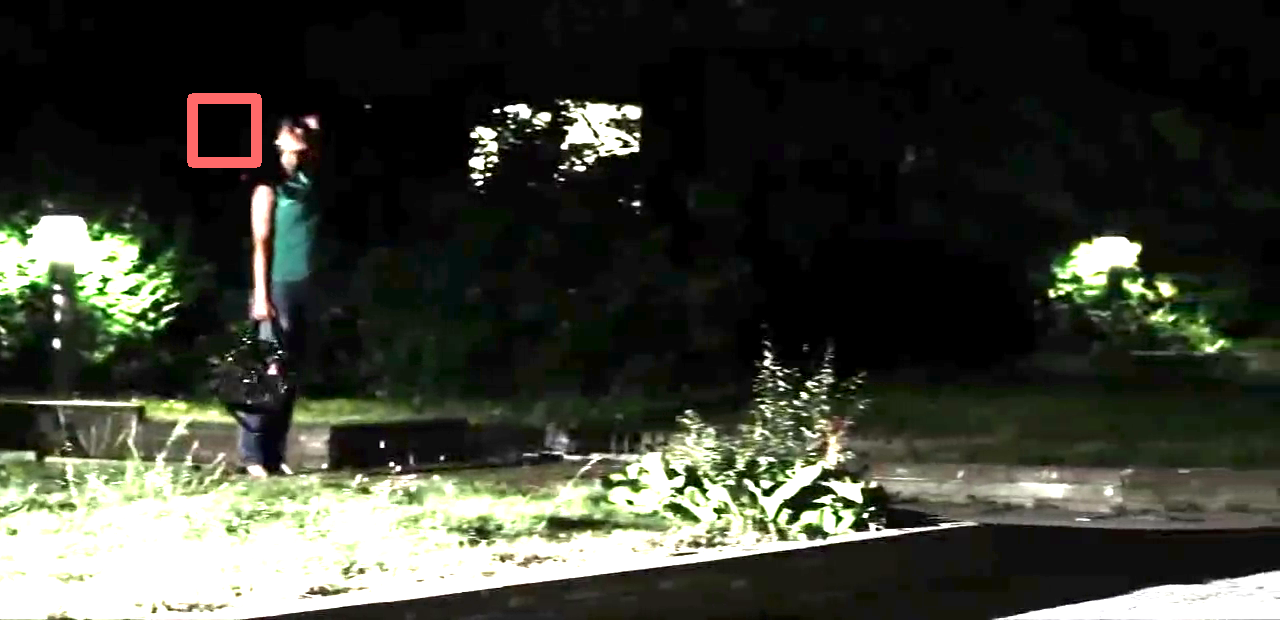}\label{fig:ava_wrong_normalised_coords}}\hspace*{1cm}
        \subfloat[Additional bounding box]{\includegraphics[height=0.15\textheight]{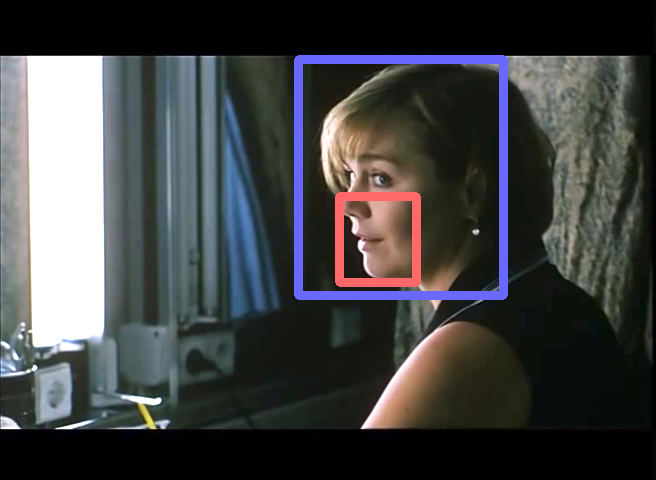}\label{fig:ava_additional_bbox}}\vspace{0.15cm}
        \captionsetup{justification=justified}
        \caption{Examples of bounding box inaccuracies in the AVA-ActiveSpeaker dataset. Correctly placed bounding boxes are indicated in blue and mistake cases in red bounding boxes.}
        \label{fig:ava_bboxes}
    \end{figure*}

    Despite being the first large-scale task-agnostic dataset published for active speaker detection, AVA-ActiveSpeaker has several drawbacks. These include a relative scarcity of records featuring speaking activity, several dubbed videos, a lack of support for pose-based active speaker detection, and unreliable face boundary coordinate information. While AVA-ActiveSpeaker provides annotations for the bounding coordinates of the face crop of a target individual, it lacks corresponding coordinates for the person's body location. Although the body position can be inferred from the scene context and head position information, the reliability of the face bounding box coordinates of AVA-ActiveSpeaker is problematic, posing challenges for accurate body localisation. Specifically, the bounding box coordinates in AVA-ActiveSpeaker are normalised to the scene dimensions, ranging from 0 to 1, based on video dimensions that do not correspond to the actual resolution of the video files in the dataset. Consequently, denormalising the face crop position information can result in bounding boxes that either exclude the target person's face entirely or include it only partially. Figure~\ref{fig:ava_bboxes} provides examples of inaccuracies in the head bounding box positions, including cases where the bounding box surrounds something that is not the head of a person, misplacements due to incorrect normalisation or other reasons, and additional bounding boxes nested within each other for the same person.

    To address the limitations of AVA-ActiveSpeaker, other datasets have been published, such as Active Speakers in the Wild (ASW)~\cite{kim2021} and Wilder Active Speaker Detection (WASD)~\cite{roxo2024}. WASD, in particular, offers a larger number of records with speaking activity and reliable annotations of both face and body boundary coordinates. Unlike AVA-ActiveSpeaker, both ASW and WASD distinguish only between cases of presence of speaking activity and absence thereof. The correlation between body motion and speech activity has long been studied~\cite{vajaria2008} and has been exploited for active speaker detection in well-controlled datasets~\cite{chakravarty2015,shahid2021}.

    \subsection{WASD Dataset Overview}\label{sec:asd-wasd}
    
        WASD stands out as the first published dataset for active speaker detection in the wild that contains reliable pose information. In this paper, we employ WASD as our benchmark due to its high diversity and the availability of reliable body boundary coordinate information.
        
        WASD was compiled from 164 YouTube videos from real-world interactions, each capped at a maximum length of 15 minutes, following a similar practice established during the creation of AVA-ActiveSpeaker. These videos were segmented into clips of up to 30 seconds, triple the length of the longest clip in AVA-ActiveSpeaker. The dataset comprises a total of 30 hours of video annotations, divided into training and validation sets in an 80/20 proportion, as established by Roxo et al.~\cite{roxo2024}.
        
        The WASD video selection aimed to ensure demographic balance across language, ethnicity, and gender. Additionally, to maintain similar demographic representation, the video quality and face resolution across both training and validation sets, the same clip can be shared by both splits. This approach allows individuals in a scene to be assigned to either split, often resulting in individuals from the same scene being assigned to different splits, which raises a notable concern for models that learn relationships between the target individual and other context individuals present in the scene. For instance, a target individual might be evaluated for their speaking activity in the validation set; however, that same individual could have already been introduced as a context individual during the training of another target individual from the same scene. This overlap can compromise the evaluation, as part of the input for validation may resemble information the model encountered during training. Nevertheless, this aspect of the dataset does not adversely affect models that focus solely on detecting the speaking activity of a target individual without considering context individuals, which is the case for the model proposed in this paper.
        
        WASD videos are categorised into five groups based on the challenge levels of visual and acoustic modalities (cf. Figure~\ref{fig:wasd_categories}). These categories include `Optimal Conditions' (OC), where there is minimal speech overlap and all speakers are fully visible without occlusion, `Speech Impairment' (SI), characterised by frequent speech overlap, `Face Occlusion' (FO), with prevalent cases of mouth occlusion, `Human Voice Noise' (HVN), featuring frequent overlap of target individuals' voices with background human noise, and `Surveillance Settings' (SS), comprising camera footage with uncertain face visibility, speech quality, or subject cooperation. Each category is balanced in terms of annotation hours and demographic representation. Roxo et al.~\cite{roxo2024} propose this categorisation as a means to assess both the adaptability of ASD models to different scenarios and the factors that are more relevant to ASD. We leverage this categorisation to evaluate scenarios where human pose information is most beneficial and those where its utilisation may not yield significant benefits.
    
    \subsection{ASD Solutions}\label{sec:asd-solutions}
    
        The publication of the AVA-ActiveSpeaker dataset marked the beginning of research on active speaker detection in the wild. Since then, various approaches have been proposed, including a baseline introduced by the authors of the dataset~\cite{roth2020}, consisting of a two-stream end-to-end neural network based on the MobileNet architecture~\cite{howard2017}. Chung~\cite{chung2019b} and Zhang et al.~\cite{zhang2019} devise approaches that surpass this baseline by employing two-stream end-to-end neural networks with 3D convolutions in their visual streams. However, these approaches are limited to short-term windows, roughly 0.5 seconds long, and rely on large-scale pre-training on lip synchronisation datasets.
        
        To overcome these issues, Alc{\'{a}}zar et al.~\cite{alcazar2020} introduce Active Speakers in Context (ASC), an ASD model that employs self-attention to infer inter-speaker relations, allowing for the consideration of long-term contexts where whole words could be pronounced, approximately 2.25 seconds. This approach addresses the limitations of large-scale pre-training on lip synchronisation datasets. However, ASC's architecture is not end-to-end; each speaker's visual and acoustic embeddings have to be provided by a previously trained short-term encoder to capture inter-speaker relationships accurately. After obtaining the embeddings, ASC stacks them as a tensor and passes them through a self-attention layer, which inferred pairwise inter-speaker relations. Subsequently, the tensor output by that layer serves as input to a temporal refinement layer, consisting of a long short-term memory (LSTM). The LSTM acts as a long-term pooling layer, refining the weighted features in the tensor by directly attending to their temporal structure. Notably, ASC was the first openly published ASD model, driving research in the area forward and inspiring direct extensions of the model~\cite{carneiro2021,kopuklu2021,sharma2023,zhang2021b}.
        
        Three major enhancements to ASC have greatly improved the performance of ASD models: firstly, the utilisation of graph neural networks for more accurate inference of inter-speaker relations~\cite{alcazar2021,alcazar2022,min2022}; secondly, the incorporation of self-attention (or transformer-based) layers to enhance the temporal modelling capabilities of the models~\cite{jiang2023,tao2021,zhang2021a,zhang2022}; thirdly, the adoption of cross-attention mechanisms to capture intermodal signals, thus enhancing the audiovisual synchronisation capabilities of the ASD models and enabling them to correlate the facial movements of a target individual with the scene audio without relying on facial information from other people in the scene~\cite{datta2022,jiang2023,jung2024,radman2024,tao2021,wang2024,werkaixi2022,xiong2023}.

        It is worth noting that extending these models to include human pose as an additional modality input would have prohibitive impacts for a couple of reasons. Firstly, many models rely on inferring inter-speaker relations for accurate localisation of speaking activity. However, in WASD, individuals in the same scene may be split between training and validation sets. Therefore, to accurately learn inter-speaker relations, data from the validation split could be included as part of the model input during training, contaminating the evaluation process. Secondly, models that leverage visual information exclusively from the target individual utilise cross-attention mechanisms to enhance the correlation between facial movements and scene audio. However, the size of these cross-attention matrices grows quadratically with the length of the scene. Since these matrices correlate modalities pairwise, the number of matrices also increases proportionally to the square of the number of modalities. Given that WASD video blocks can be more than three times longer than the longest video block in AVA-ActiveSpeaker, employing architectures with cross-attention mechanisms becomes impractical, especially for embedded applications. Given these constraints, we chose to build upon Light-ASD~\cite{liao2023}, the only high-performing ASD model at the time of writing that does not employ cross-attention mechanisms and relies exclusively on visual information from the target speaker to detect speech activity.

\section{Light-ASD Model Architecture}\label{sec:lightasd}

    Instead of relying on complex models with high memory and computational requirements, Light-ASD prioritises resource efficiency while maintaining competitive performance. This is accomplished through various strategies, including simplified feature extraction, employing bidirectional gated recurrent units (BiGRUs)~\cite{cho2014} for cross-modal modelling, and optimising the model architecture for efficiency~\cite{liao2023}. Notably, Light-ASD achieves comparable results to state-of-the-art methods while significantly reducing model parameters and floating-point operations (FLOPs). Its modular design and reduced resource requirements make it easily extendable to incorporate additional modalities and well suited for deployment in resource-constrained environments.

    To minimise computational burden, Light-ASD does not leverage the relational contextual information between speakers. Instead, it relies solely on the information of a single target candidate to accurately detect instances of speech activity. Additionally, it is an end-to-end model that does not require pre-training on external training data, resulting in reduced processing time. The model receives a sequence of facial crops of the target individual and the corresponding audio from the video clip as input. The face and audio inputs undergo separate processing by feature encoders specific to each modality, with each encoder producing a feature tensor. These tensors are then fed into a detection module, which assigns a score to each frame, indicating the likelihood of the target individual being actively speaking.
    
    \subsection{Feature Encoders}\label{sec:lightasd-feature}
    
        Light-ASD's face feature encoder\footnote{Liao et al.~\cite{liao2023} use the term ``visual feature encoder.'' Here, we replace it with ``face feature encoder'' to distinguish the processing of facial information from other visual cues, such as body pose. Further terminologies associated with facial information are adjusted accordingly.} processes 3D stacks of greyscale face images, each with uniform height and width dimensions denoted by $N_{f}$. In contrast, the audio feature encoder handles 2D maps consisting of sequences of vectors containing $N_{a} = 13$ mel-frequency cepstrum coefficients (MFCCs). The length of the image stack represents the number of frames and is denoted by $T_{f}$, while that of the MFCC sequence is given by $T_{a} = 4 \, T_{f}$. To ensure alignment, the raw audio of the video clip is converted to MFCCs, and the sequence of MFCC vectors is padded or truncated to match exactly four times the number of frames. Additionally, both face and audio inputs have a channel dimension, denoted as $C_{in}$, set to $1$ for each modality due to the greyscale nature of the images and the single-channel representation of MFCCs. Consequently, the input to the face feature encoder is a tensor of dimensions $1 \times N_{f} \times N_{f} \times T_{f}$, whereas the input to the audio feature encoder is a tensor of dimensions $1 \times N_{a} \times T_{a}$. For simplicity, the image height and width dimensions as well as the MFCC vector dimension are hereafter referred to as spatial dimensions, while those associated with the image stack and the MFCC sequence are termed temporal dimensions.
        
        \begin{figure*}[!ht]
            \centering
            \captionsetup{justification=centering}
            \subfloat[Face feature encoder]{\includegraphics[height=0.135\textheight]{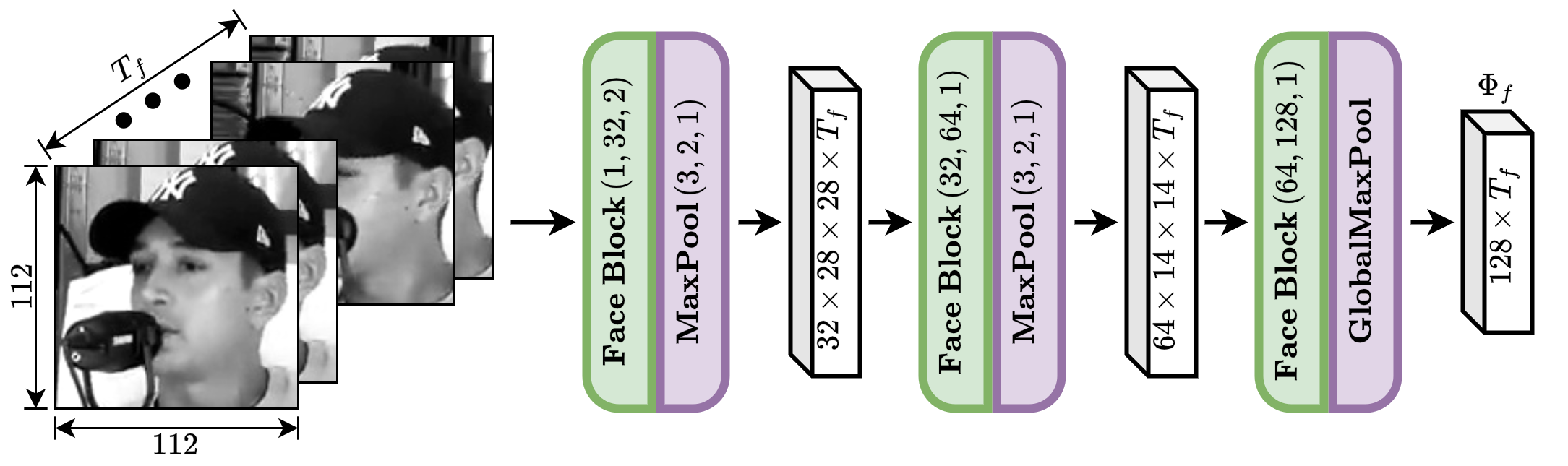}\label{fig:lightasd-face_feature_encoder}}\\
            \subfloat[Audio feature encoder]{\includegraphics[height=0.135\textheight]{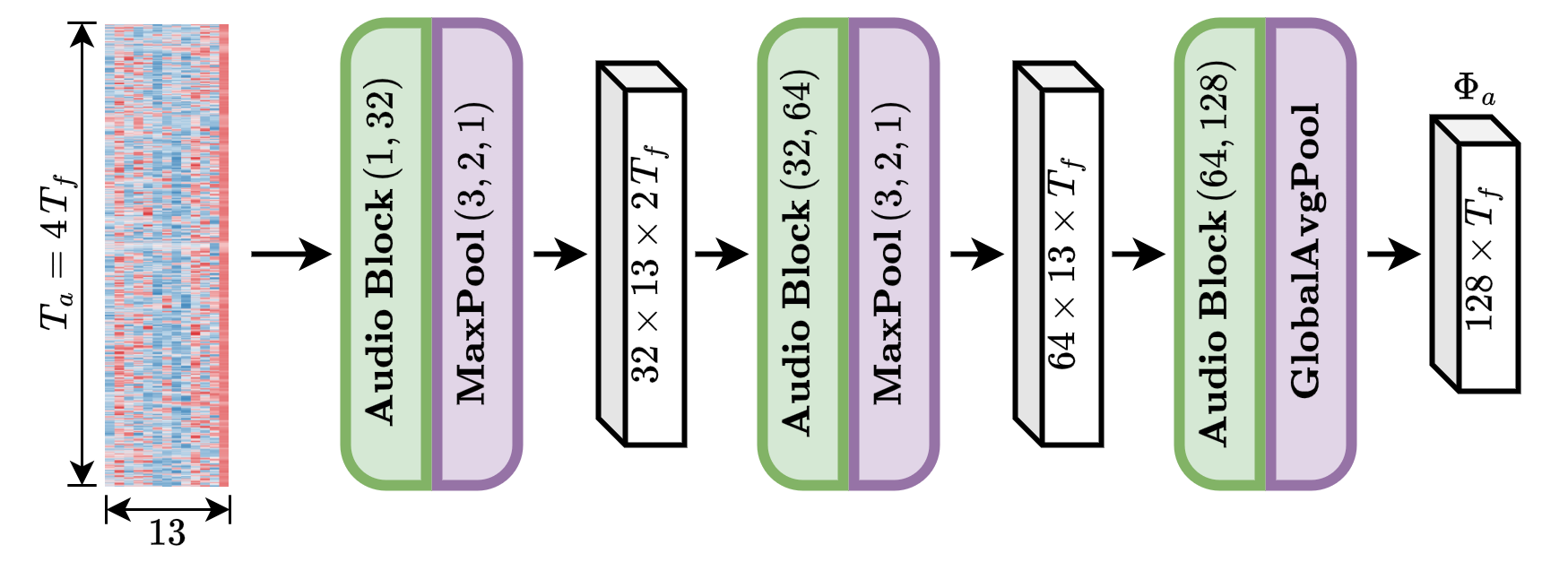}\label{fig:ligtasd-audio_feature_encoder}}\\
            \subfloat[Face block]{\includegraphics[height=0.385\textheight]{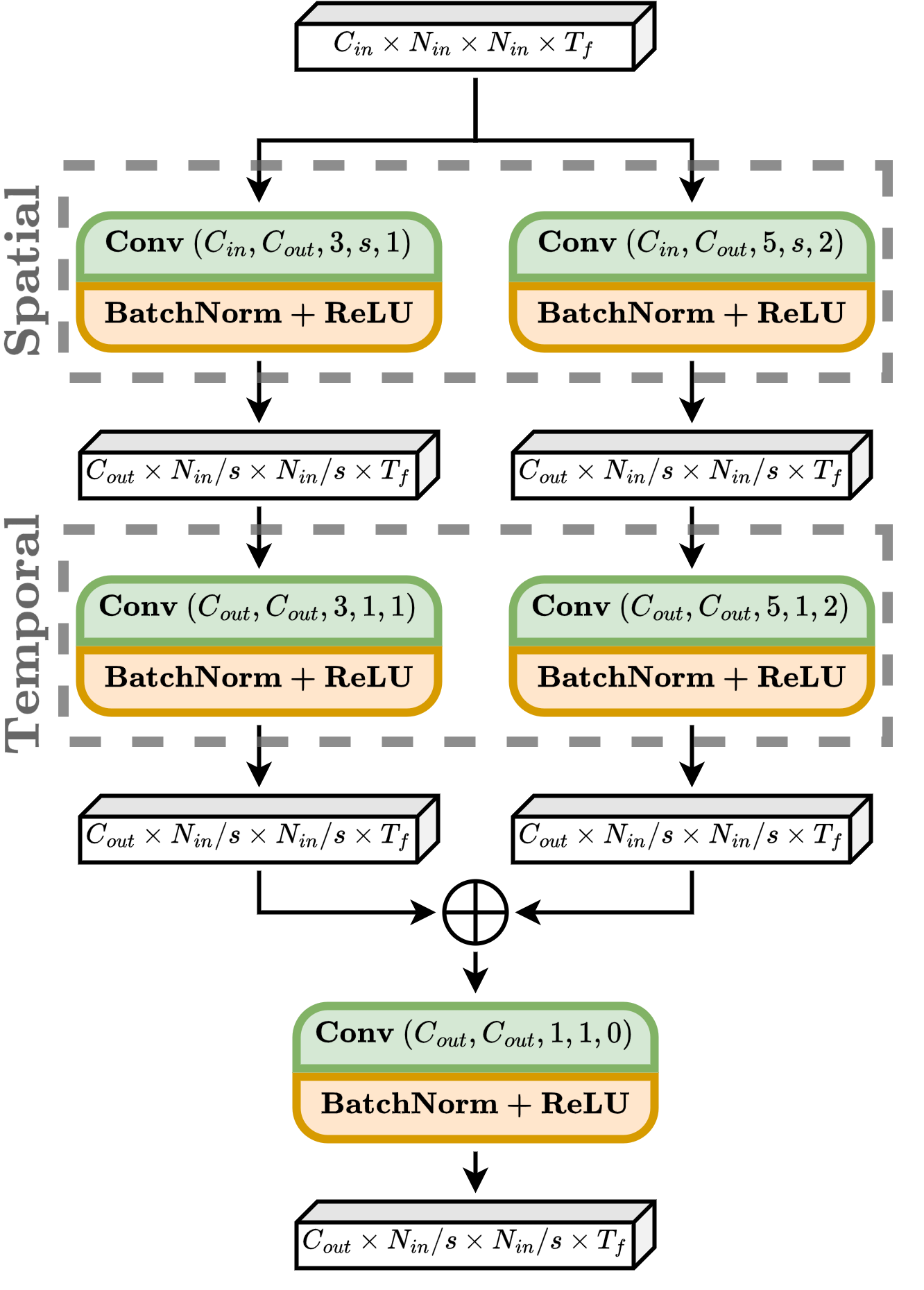}\label{fig:lightasd-face_block}}\hfil
            \subfloat[Audio block]{\includegraphics[height=0.385\textheight]{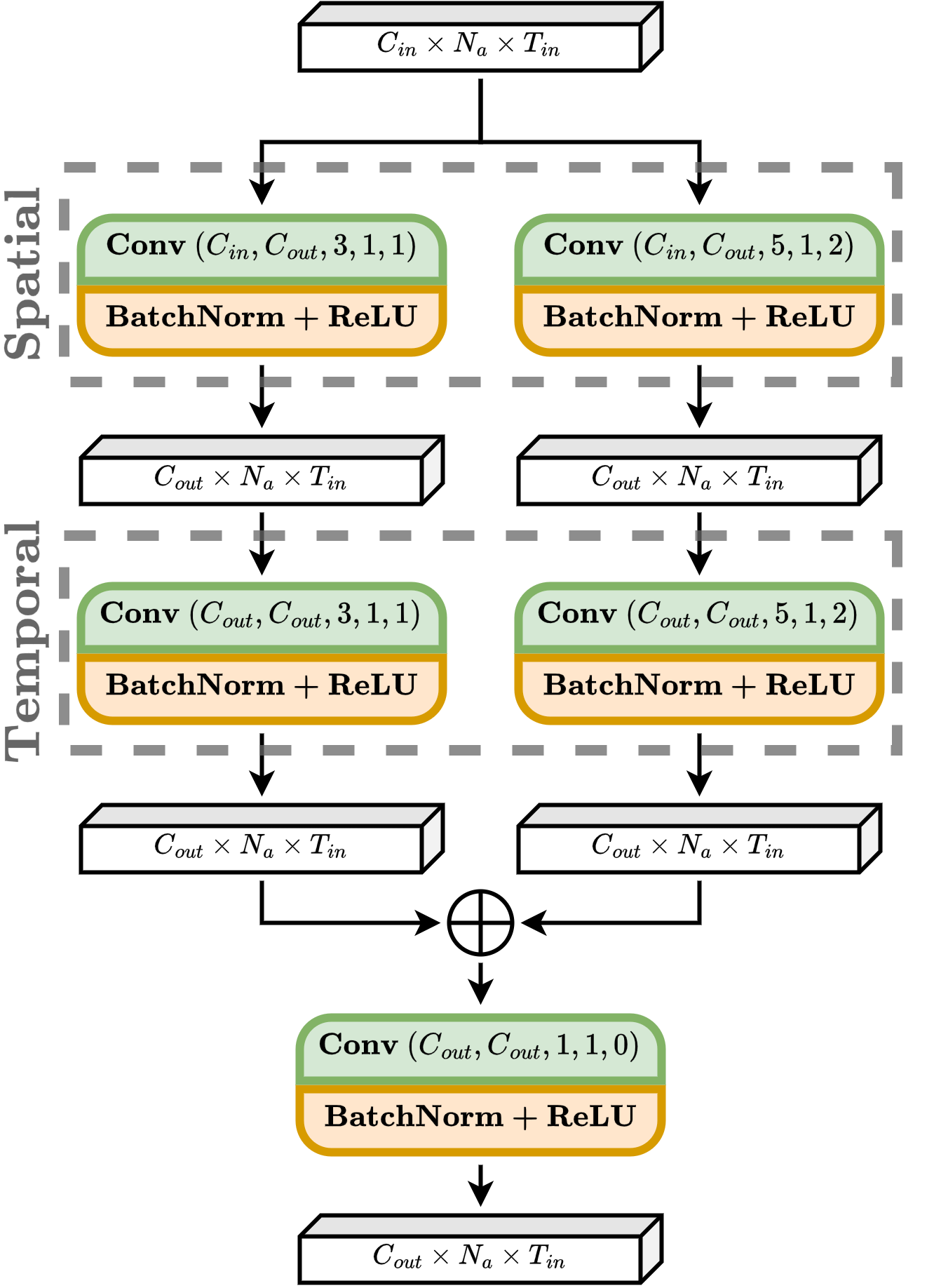}\label{fig:ligtasd-audio_block}}\vspace{0.15cm}
            \captionsetup{justification=justified}
            \caption{Architectures of Light-ASD feature encoders and their inner blocks. Blocks are given by their number of in-channels $C_{in}$ and out-channels $C_{out}$, and in the face feature encoder, the stride $s$ in both spatial dimensions. Pooling layer parameters are the kernel size, stride, and padding size. In the face feature encoder, pooling is applied on both spatial dimensions, while in the audio feature encoder, it is applied on the temporal dimension. The parameters of the convolution layers within the blocks are the number of in-channels, number of out-channels, kernel size, stride, and padding size. The labels beside each convolution layer indicate whether the convolution is applied on the spatial or temporal dimensions.}
            \label{fig:lightasd-featureencoders}
        \end{figure*}
        
        Both feature encoders of Light-ASD share similar architectures, as illustrated in Figures~\ref{fig:lightasd-face_feature_encoder} and~\ref{fig:ligtasd-audio_feature_encoder}. Each encoder comprises three modality-specific blocks followed by pooling layers. Due to the relation between $T_{a}$ and $T_{v}$, and to ensure identical dimensions for both face and audio features, the first two pooling layers of the audio feature encoder perform a 1D max pool dimensionality reduction on the temporal dimension, while the corresponding layers of the face feature encoder perform a 2D max pool dimensionality reduction operation on the spatial dimensions. Furthermore, in both encoders, the last pooling layer operates on the spatial dimensions; however, a global max pool is performed in the face feature encoder, while a global average pool is utilised in the audio feature encoder.
        
        The architectural similarity also extends to the block level, as depicted in Figures~\ref{fig:lightasd-face_block} and~\ref{fig:ligtasd-audio_block}. Each modality-specific block includes two paths of feature extraction with convolutions of distinct kernel sizes ($\kappa = 3$ and $\kappa = 5$). Instead of using a single high-dimensional convolution that covers all tensor dimensions, the convolution is split into two sequential convolutions, with the first one operating along the spatial dimensions and the next along the temporal dimension. This split in the convolution process significantly reduces the number of parameters in the model. Both spatial and temporal convolutions in either path have the same kernel size. The tensors are padded accordingly to maintain their dimensions after a convolution operation. Batch normalisation~\cite{ioffe2015} and ReLU activation are performed after each convolution. The presence of multiple feature extraction paths ensures a variety of representations, which are then integrated by summing the representations and applying a convolution with a kernel size of $1$. All convolutions have a stride of $1$ except for the spatial convolutions in both paths of the first block of the face feature encoder, which has a stride of $2$. Finally, the number of input and output channels of each block are aligned, with both feature encoders having $1$ in-channel and $32$ out-channels in their first block, $32$ in-channels and $64$ out-channels in their second block, and $64$ in-channels and $128$ out-channels in their last block. The aligned architectures of the feature encoders allow them to output face and audio features $\Phi_{f}$ and $\Phi_{a}$ with identical dimensions, namely $128 \times T_{f}$.

    \subsection{Modality Fusion and Prediction}\label{sec:lightasd-fusion}

        The fusion of the face and audio features $\Phi_{f}$ and $\Phi_{a}$ is achieved through an element-wise sum. Subsequently, a BiGRU is employed to capture the temporal context inherent in the resulting multimodal representation $\Phi_{fa}$. Leveraging this temporal context, a fully connected (FC) layer provides two scores for every frame of the video clip, indicating the likelihood of the target individual actively speaking or not in that frame. From these scores, Light-ASD performs the prediction using a softmax function.

        Auxiliary classification heads are also utilised for training purposes in Light-ASD. These heads share a similar structure, comprising a BiGRU and a fully connected layer. However, unlike the main classification head, which integrates all modalities, each auxiliary classification head operates within a purely unimodal framework. Therefore, in each auxiliary classification head, the features output by the feature encoders are directly fed into the corresponding BiGRU.
        
        Light-ASD includes one auxiliary classification head for each auxiliary loss. Notably, while the face auxiliary classifier can determine if a target individual is speaking solely based on facial information, the audio auxiliary classifier can only determine if someone is speaking overall when no facial cues are provided, resulting in high losses. To address this issue, Light-ASD incorporates both the main and face auxiliary losses in its training, excluding the audio auxiliary loss.
        
        For the computation of the loss function, a temperature parameter $\tau$ is utilised to adjust the calculation of the probability of the target individual being actively speaking. The probability of the target individual being actively speaking at the $t$\textsuperscript{th} frame is determined by
        \begin{equation}
            p^{t}_{M} = \dfrac{\mathrm{exp} \! \left( \sigma^{t}_{M, spk} / \tau \right)}{\mathrm{exp} \! \left( \sigma^{t}_{M, spk} / \tau \right) + \mathrm{exp} \! \left( \sigma^{t}_{M, sil} / \tau \right)},
            \label{eq:lightasd-prediction}
        \end{equation}
        where $M$ represents the modalities used in the classification head, and $\sigma^{i}_{M, spk}$ and $\sigma^{i}_{M, sil}$ are the scores assigned by that classification head to the likelihood of the target individual being actively speaking or silent respectively. The temperature $\tau$ progressively decreases with each epoch, inversely correlating with the epoch number, following the formula
        \begin{equation}
            \tau = 1.3 - 0.02 \xi,
            \label{eq:lightasd-temperaturedecay}
        \end{equation}
        where $\xi$ is the epoch number. The temperature adjustment facilitates the model's exploration of the solution space, aiding in the avoidance of local optima and guiding the model towards a more refined solution.
    
        \begin{figure*}[!ht]
            \centering
            \captionsetup{justification=justified}
            \includegraphics[height=0.15\textheight]{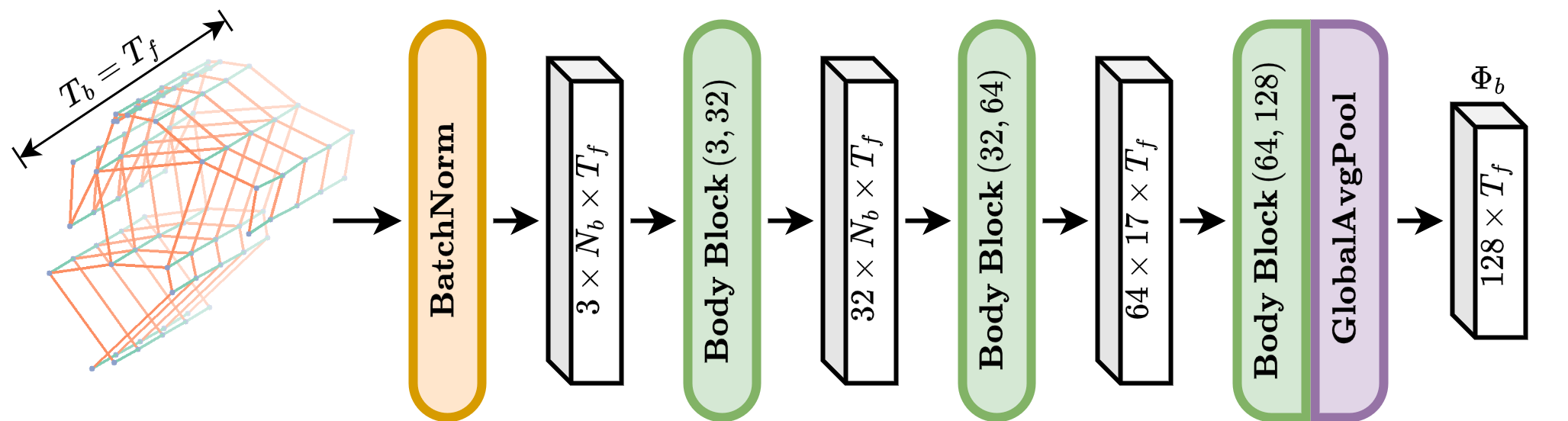}\vspace{0.15cm}
            \caption{FabuLight-ASD's body encoder. The parameters of each body block indicate the number of in- and out-channels.}
            \label{fig:fabulightasd-bodyfeatureencoder}
        \end{figure*}

        Given a video clip with $T$ frames or a batch thereof, one loss $\mathcal{L}_{M}$ is calculated for each classification head. Light-ASD utilises the main multimodal loss $\mathcal{L}_{f a}$ and the face auxiliary loss $\mathcal{L}_{f}$. The loss $\mathcal{L}_{M}$ is a cross-entropy loss, calculated as follows
        \begin{equation}
            \mathcal{L}_{M} = \dfrac{1}{T} \sum\limits_{t = 1}^{T} \left( g^{t} \mathrm{log} \! \left( p^{t}_{M} \right) + \left( 1 - g^{t} \right) \mathrm{log} \! \left( 1 - p^{t}_{M} \right) \right),
            \label{eq:lightasd-modalityloss}
        \end{equation}
        where $g^{t}$ is the ground truth about the status of the target individual as an active speaker at the $t$\textsuperscript{th} frame, with $g^{t} = 1$ indicating presence of active speech and $g^{t} = 0$ indicating absence thereof. The total loss is given by
        \begin{equation}
            \mathcal{L}_{\mathrm{total}} = \mathcal{L}_{f a} + 0.5 \mathcal{L}_{f}.
            \label{eq:lightasd-totalloss}
        \end{equation}

        For evaluation purposes, the temperature is set to a fixed value of $\tau = 1$, and only the probability provided by the multimodal classification is used to determine the presence or absence of active speech. The variation in the value of $\tau$ and the probabilities output by the auxiliary classification heads are utilised exclusively during training.

\section{FabuLight-ASD Model Architecture}\label{sec:fabulightasd}

    FabuLight-ASD harnesses body pose information to enhance the inference of whether a target individual is actively speaking. This capability proves particularly valuable in scenarios where the person is distant from the camera, resulting in a low resolution of their facial features.
    
    \subsection{Body Pose Stream}\label{sec:fabulightasd-pose}
    
        Alongside the face and audio feature encoders, FabuLight-ASD incorporates a body feature encoder. The body pose information is represented as a set of $N_{b}$ body joints in the COCO format~\cite{lin2014}\footnote{We consider two possible ways of feeding the encoder with the information of the body pose of a target individual:
        \begin{inparaenum}[(\itshape i\upshape)]
            \item the set of body joints of the whole body, which encompasses $N_{b} = 17$ joints in the COCO format; or
            \item the set of body joints of the upper body, comprising $N_{b} = 11$ joints. 
        \end{inparaenum}}. Given the body bounding box coordinates provided in WASD, the HRNet~\cite{sun2019} implementation within MMPose~\cite{mmpose2020} is utilised to infer the set of joints corresponding to the person's body delimited by the bounding box coordinates. Each joint is characterised by three values: the horizontal position, the vertical position, and a confidence score in the range $\left[ 0, 1 \right]$ indicating HRNet's confidence in the provided positions.
            
        \subsubsection{Body Feature Encoder}\label{sec:fabulightasd-bodyfeatureencoder}

            Drawing parallels with the architecture of Light-ASD, FabuLight-ASD's body feature encoder incorporates three modality-processing blocks, as presented in Figure~\ref{fig:fabulightasd-bodyfeatureencoder}. To match the configurations of the face and audio feature encoders inherited from Light-ASD, the number of in-channels and out-channels of each body block in FabuLight-ASD corresponds to those in the other feature encoders, except the in-channels of the first block, which must align with the number of channels of the model input. This input comprises three channels representing each body joint's horizontal position, vertical position, and the confidence of the pose estimation model regarding those coordinates. Accordingly, FabuLight-ASD's body blocks, from the first to the last, have $3$ in-channels and $32$ out-channels, $32$ in-channels and $64$ out-channels, and $64$ in-channels and $128$ out-channels.
            
            To condense the feature representations of each skeleton, a global average pooling operation is applied across the spatial dimension, thus reducing the initial $128 \times N_{b}$ feature representation to a single $128$-dimensional feature vector for each skeleton. Consequently, the body feature encoder processes a sequence of human body pose skeletons with dimensions $3 \times N_{b} \times T_{f}$ as input, generating a body feature representation $\Phi_{b}$ whose dimensions align to those of $\Phi_{f}$ and $\Phi_{a}$, specifically $128 \times T_{f}$.
        
        \subsubsection{Body Block Architecture}\label{sec:fabulightasd-bodyblock}

            The body blocks within the encoder are based on the spatial-temporal graph convolutional networks (ST-GCNs) proposed by Yan et al.~\cite{yan2018}. This choice is motivated by three key factors. Firstly, ST-GCNs excel at capturing dynamic patterns and interactions among body parts during actions. By utilising them as the backbone of its body feature encoder, FabuLight-ASD can discern relationships between body joints, extracting relevant cues about a target individual's speaking activity. Secondly, although initially designed for action recognition, ST-GCNs have proven versatile and their core concept has been applied in other tasks, such as active speaker detection~\cite{alcazar2021,alcazar2022}, where dynamic relationships among individuals are represented using graph convolutional networks (GCNs). In these scenarios, individuals are depicted as graph nodes, with edges illustrating their dynamic interactions. Thirdly, ST-GCNs inherently leverage both spatial and temporal information to model the relations between body joints. This aligns with the architecture of the convolutional networks used in the modality-specific blocks of Light-ASD feature encoders, indicating the feasibility of adapting ST-GCNs to maintain the lightweight nature of the architecture while effectively capturing the dynamics of body pose information.
            
\begin{figure}[!ht]
                \centering
                \captionsetup{justification=justified}
                \includegraphics[height=0.3\textheight]{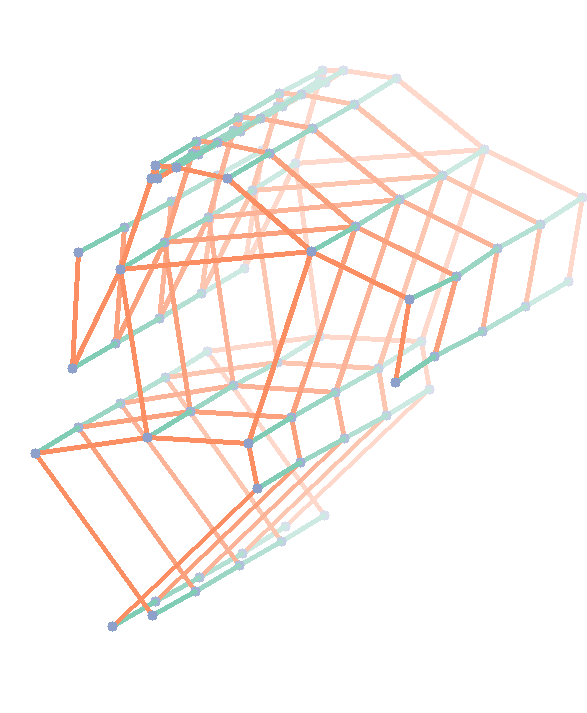}\vspace{0.15cm}
                \caption{Spatial-temporal graph representing a sequence of body poses. Body joints are marked in blue. Edges indicating joint connections within the same frame are coloured in orange. Temporal connections, which link each joint in a given frame to the corresponding joints in adjacent frames, are represented in green. Differing alpha channels are used to indicate body pose skeletons in different frames and for visualisation purposes.}
                \label{fig:fabulightasd-spatialtemporalgraph}
            \end{figure}

            The poses of a person across a sequence of frames are represented as a sequence of skeletons, each with consistently numbered joints. This sequence of skeletons is represented as a graph $G = \left( V , E \right)$, where $V$ denotes the joints of the skeleton sequence, with $v_{t, i} \in V$ representing the $i$\textsuperscript{th} joint at the $t$\textsuperscript{th} frame. The edge set $E$ comprises two subsets: $E_{S}$, depicting joint connections within each frame following the COCO format, and $E_{F}$, linking each joint in a frame to corresponding joints in adjacent frames. Formally, $v_{t, i}$ is connected to both $v_{t - 1, i}$ and $v_{t + 1, i}$. Figure~\ref{fig:fabulightasd-spatialtemporalgraph} illustrates the spatial-temporal graph of a skeleton sequence.

\begin{figure}[!th]
                \centering
                \captionsetup{justification=justified}
                \includegraphics[height=0.425\textheight]{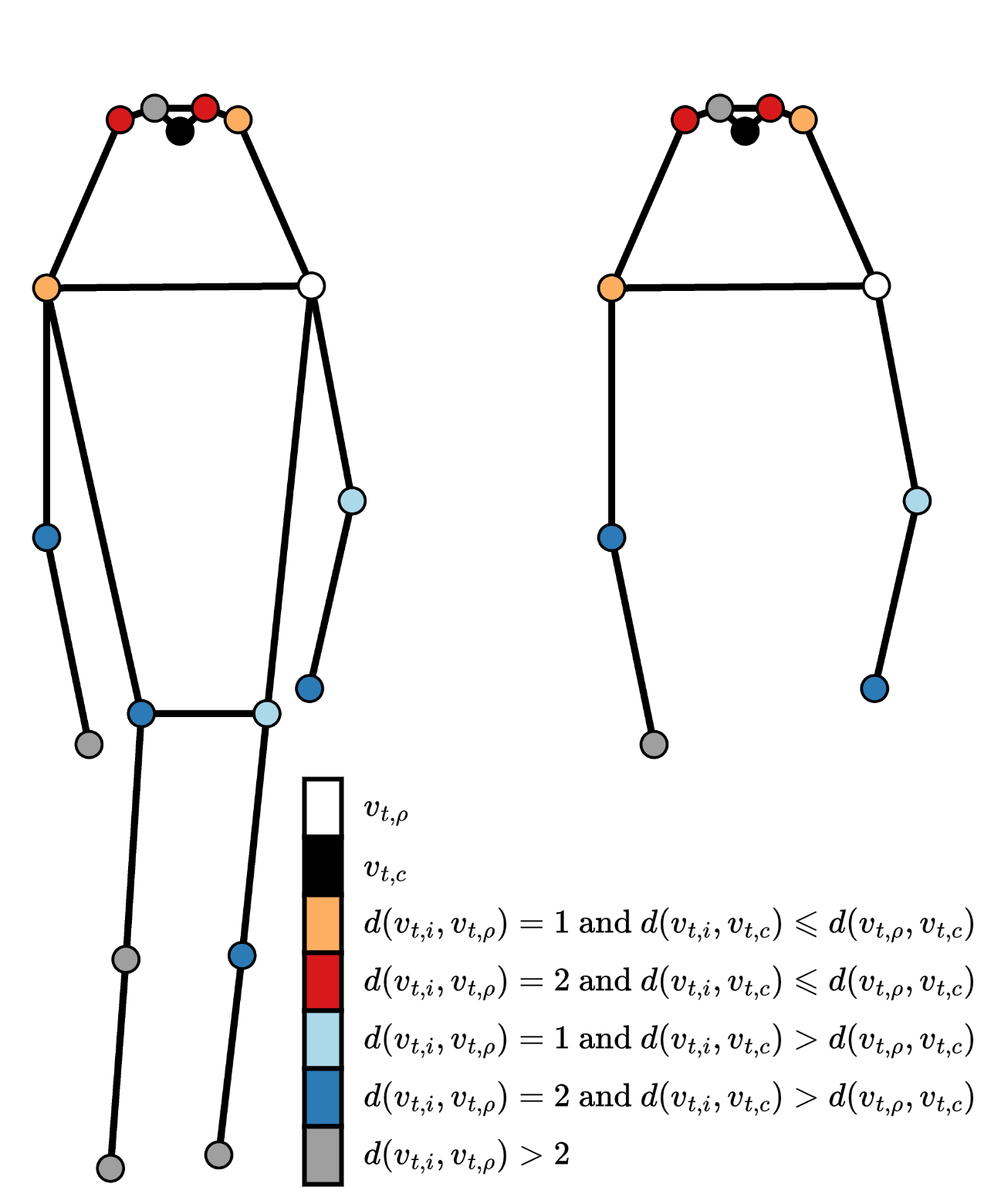}\vspace{0.15cm}
                \caption{Spatial-configuration-based node partition of the body pose skeleton graph using the COCO pose template. The left side illustrates the partitioning for the whole body, while the right side focuses on the upper body. For the COCO template, ST-GCN treats the nose node as the central node $v_{t,c}$ for determining the spatial configuration. The spatial configuration is the same for every time frame $t$. In this example, the right shoulder node is taken as the root node $v_{t,\rho}$. Nodes $v_{t,i}$ are marked based on their distance from the root node and whether they lie closer or further from $v_{t,c}$ than the root node. This example represents the case in which $R = 2$ and the spatial kernel size is $\kappa_{S} = 5$. Given this configuration, the adjacency matrix is determined according to Equation~\ref{eq:fabulightasd-adjacentmatrix}.}
                \label{fig:fabulightasd-spatialpartition}
            \end{figure}

            The body blocks generate increasingly higher-level feature maps on the graph. The convolutions employed in these blocks consider the neighbourhood of each graph node, encompassing intra-body connections and self-connections represented through an adjacency matrix. The concept of node neighbourhood is extended to include nodes connected not only within the same frame, denoted $E_{S}$, but also across adjacent frames, denoted $E_{F}$. Additionally, the spatial configuration partition strategy divides the spatial neighbour set based on the distance between nodes and their distances to the skeleton's central node $v_{t, c}$ at frame $t$. This partition distinguishes between concentric and eccentric body part motions. It is represented by $2R + 1$ adjacency matrices $\mathbf{A}^{-R}, \cdots, \mathbf{A}^{R}$, where $R$ is the maximum distance threshold for considering node connections. Specifically, nodes are included in an adjacency matrix if they lie within at most $R$ units from each other. Each adjacency matrix $\mathbf{A}^{r} = {\left[ A^{r}_{i j} \right]}_{N_{b} \times N_{b}}$ is defined as
            \begin{equation}
                A^{r}_{i j} = \left\{ \begin{array}{rl}
                                          1, & \text{if } r = 0 \text{ and } i = j \\[.125cm]
                                          1, & \text{if } r < 0, \, d \! \left( v_{t, i} , v_{t, j} \right) = r, \\
                                          & \text{and } d \! \left( v_{t, i} , v_{t, c} \right) \leqslant d \! \left( v_{t, j} , v_{t, c} \right) \\[.125cm]
                                          1, & \text{if } r > 0, \, d \! \left( v_{t, i} , v_{t, j} \right) = r, \\
                                          & \text{and } d \! \left( v_{t, i} , v_{t, c} \right) > d \! \left( v_{t, j} , v_{t, c} \right) \\[.125cm]
                                          0, & \text{otherwise} \\
                                       \end{array} \right. .
                \label{eq:fabulightasd-adjacentmatrix}
            \end{equation}
            Here, $d(\cdot, \cdot)$ represents the smallest distance between two nodes. Figure~\ref{fig:fabulightasd-spatialpartition} illustrates the labelling of the skeleton graph nodes according to the spatial configuration partition.
            
            Each body block takes as input a tensor $\mathbf{X} \in \mathbb{R}^{C_{in} \times N_{b} \times T_{in}}$ and produces a tensor $\mathbf{Y} \in \mathbb{R}^{C_{out} \times N_{b} \times T_{out}}$. Each block is associated with a spatial kernel size $\kappa_{S}$ and a temporal kernel size $\kappa_{T}$. The former is set at $\kappa_{S} = 2 R + 1$, representing the number of spatial configuration partitions, while the latter, although unrelated to the adjacency matrices, is set to be identical to $\kappa_{S}$. Akin to Light-ASD's modality-specific blocks, FabuLight-ASD's body blocks incorporate two paths of feature extraction: one utilising kernel sizes $\kappa_S = \kappa_T = 3$, and the other with $\kappa_S = \kappa_T = 5$. The determination of $\kappa_S$ for each path involves setting $R = 1$ for the former and $R = 2$ for the latter, guided by the relation $\kappa_S = 2 R + 1$, which defines the spatial kernel size based on the number of spatial configuration partitions.

            \begin{figure}[!th]
                \centering
                \includegraphics[height=0.4825\textheight]{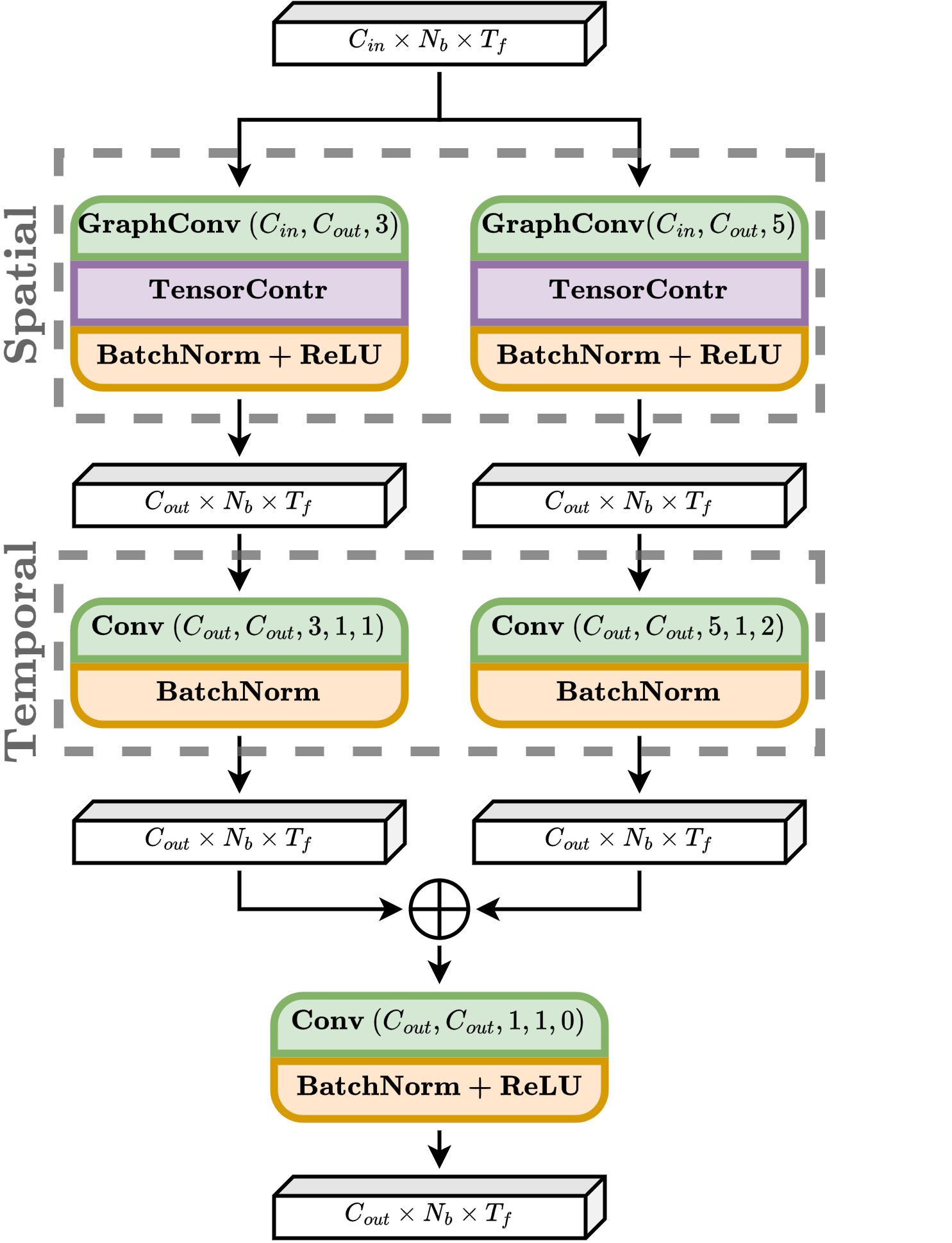}\vspace{0.15cm}
                \captionsetup{justification=justified}
                \caption{Architecture of FabuLight-ASD's body block. The parameters of each component follow the description given in Figure~\ref{fig:lightasd-featureencoders}. TensorContr stands for tensor contraction, representing the operation described by Equation~\ref{eq:fabulightasd-tensorcontraction}.}
                \label{fig:fabulightasd-bodyblock}
            \end{figure}
            
            Both paths comprise a graph convolution followed by a temporal convolution, as illustrated in Figure~\ref{fig:fabulightasd-bodyblock}. The graph convolution over an input $\mathbf{X} \in \mathbb{R}^{C_{in} \times N_{b} \times T_{in}}$ occurs in two stages: Firstly, a 2D convolution yields $\mathbf{M} \in \mathbb{R}^{\kappa_{S} \times C_{out} \times N_{b} \times T_{in}}$ using a learnable weight tensor $\mathbf{W}$ with a kernel of size $1$. Next, via a tensor contraction operation, the spatial-temporal graph convolution generates a feature tensor $\mathbf{Z} \in \mathbb{R}^{C_{out} \times N_{b} \times T_{in}}$ whose channel-wise slices $\mathbf{Z}_{c} \in \mathbb{R}^{N_{b} \times T_{in}}$ are given by
            \begin{equation}
                \mathbf{Z}_{c} = \sum\limits_{r = -R}^{R} \! {\left( \mathbf{B}^{r} \right)}^{\top} \mathbf{M}^{r}_{c},
                \label{eq:fabulightasd-tensorcontraction}
            \end{equation}
            where $\mathbf{M}^{r}_{c} \in \mathbb{R}^{N_{b} \times T_{in}}$ represents a slice of $\mathbf{M}$, and $\mathbf{B}^{r} \in \mathbb{R}^{N_{b} \times N_{b}}$ is a learnable matrix, which is initialised according to
            \begin{equation}
                \mathbf{B}^{r} = {\left( \mathbf{D}^{r} \right)}^{-\frac{1}{2}} \mathbf{A}^{r} {\left( \mathbf{D}^{r} \right)}^{-\frac{1}{2}},
                \label{eq:fabulightasd-normalisedadjacencymatrix}
            \end{equation}
            where $\mathbf{A}^{r}$ is the adjacency matrix defined by Equation~\ref{eq:fabulightasd-adjacentmatrix}, and $\mathbf{D}^{r}$ is a diagonal normalisation matrix, with $D^{r}_{i i} = \sum_{k} \left( A^{r}_{i k} \right) + \varepsilon$. Here, $\varepsilon$ is set to $0.001$ to prevent empty rows in $\mathbf{D}^{r}$. Subsequently, the temporal convolution entails a 2D convolution with kernel dimension $\kappa_{T} \times 1$ over $\mathbf{Z}$, resulting in a feature representation $\mathbf{Y} \in \mathbb{R}^{C_{out} \times N_{b} \times T_{out}}$. The feature representations produced by both paths are then integrated via summation followed by a convolution with a kernel size of $1$. Notice that each learnable parameter described in this paragraph has a copy in every path of all body blocks of FabuLight-ASD, which are initialised identically but fine-tuned differently.
            
            Batch normalisation is applied after each convolution operation and as the initial operation on the input data, before it passes through the first body block. The ReLU activation function is applied after the batch normalisation following the graph convolution and at the end of each block.

    \subsection{Modality Fusion and Prediction}\label{sec:fabulightasd-fusion}

        Similarly to Light-ASD, FabuLight-ASD features one main classification head and some auxiliary classification heads, which are exclusively used for training purposes. In Light-ASD, the main classification head generates scores based on the multimodal representation obtained by fusing the features produced by both feature encoders, namely $\Phi_{f}$ and $\Phi_{a}$. Additionally, Light-ASD has a single auxiliary classification head, responsible for generating scores solely based on the face feature $\Phi_{f}$. In contrast, FabuLight-ASD processes three input modalities: face (as a sequence of face crops of the target individual), audio (as an MFCC tensor derived from the audio clip), and body pose (as a spatio-temporal graph of the body joints of the target individual), allowing fusion by combining them. The main classification head of FabuLight-ASD generates scores based on the multimodal representation provided by fusing all modality features, namely $\Phi_{f}$, $\Phi_{a}$, and $\Phi_{b}$. Furthermore, FabuLight-ASD includes two auxiliary classification heads that utilise unimodal features, specifically face and body. FabuLight-ASD lacks an audio auxiliary classification head for the same reason as Light-ASD, as discussed in Section~\ref{sec:lightasd-fusion}.
        
        Given a set of modalities $M$, a classification head provides two scores, $\sigma^{t}_{M, spk}$ and $\sigma^{t}_{M, sil}$, for every frame $t$ of an input video footage, indicating the likelihood of a target individual being actively speaking or not in each frame respectively. Akin to Light-ASD's classification procedure, unimodal features are summed element-wise. Subsequently, a BiGRU is applied to the resulting sum, followed by an FC layer that outputs the scores. Both the main classification head and the auxiliary ones can predict the speaking activity of an individual across a sequence of frames using Equation~\ref{eq:lightasd-prediction}. The predictions of all classification heads and the temperature parameter $\tau$ are used exclusively during training. However, during evaluation, only the prediction obtained from the main classification head is employed, and no temperature adjustment is applied.
    
        \begin{table*}[htp!]
            \captionsetup{justification=justified}
            \caption{Comparison of performance on the WASD dataset for various ASD models, as obtained by Roxo et al.~\cite{roxo2024}, and for two variants of FabuLight-ASD. Values represent the mean average precision (mAP) of each model across WASD categories: OC (optimal conditions), SI (speech impairment), FO (face occlusion), HVN (human voice noise), and SS (surveillance settings). Citations beside model names refer to the papers where the models were introduced. All results were obtained by Roxo et al.~\cite{roxo2024}. The highest mAPs within each category are marked in bold. Italicised values indicate mAPs higher than those of both FabuLight-ASD variants.}
            \centering
            \tablebodyfont
            \begin{tabular}{lcccccc}
                \toprule
                Model & OC & SI & FO & HVN & SS & Overall \\
                \midrule
                ASC~\cite{alcazar2020} & 91.2 & 92.3 & 87.1 & 66.8 & 72.2 & 85.7 \\
                MAAS~\cite{alcazar2021} & 90.7 & 92.6 & 87.0 & 67.0 & 76.5 & 86.4 \\
                ASDNet~\cite{kopuklu2021} & 96.5 & 97.4 & 92.1 & 77.4 & \textit{77.8} & 92.0 \\
                TalkNet (face + audio)~\cite{tao2021} & 95.8 & 97.5 & 93.1 & 81.4 & \textit{77.5} & 92.3 \\
                TS-TalkNet~\cite{jiang2023} & 96.8 & 97.9 & 94.4 & 84.0 & \textit{79.3} & 93.1 \\
                Light-ASD~\cite{liao2023} & 97.8 & 98.3 & 95.4 & 84.7 & \textit{77.9} & 93.7 \\
                TalkNet (body + audio)~\cite{roxo2024} & 91.1 & 95.5 & 88.4 & 73.1 & 75.0 & --- \\
                TalkNet (face + body + audio)~\cite{roxo2024} & 96.9 & 98.1 & 95.4 & 83.8 & \textit{\textbf{81.5}} & --- \\
                \midrule
                \textbf{FabuLight-ASD (upper body)} & 97.7 & \textbf{98.6} & \textbf{96.1} & \textbf{86.4} & 77.3 & \textbf{94.3} \\
                \textbf{FabuLight-ASD (whole body)} & \textbf{98.1} & \textbf{98.6} & 96.0 & 85.6 & 77.1 & 94.0 \\
                \bottomrule
            \end{tabular}
            \label{tab:experiments-comparisonfabulight}
        \end{table*}
        
        Given the probabilities $p^{t}_{fab}$, $p^{t}_{f}$, and $p^{t}_{b}$, computed from the scores output by the classification heads, the corresponding cross-entropy losses are calculated using Equation~\ref{eq:lightasd-modalityloss}. The total loss of the model is then determined by
        \begin{equation}
            \mathcal{L}_{\mathrm{total}} = \mathcal{L}_{f a b} + 0.25 \mathcal{L}_{f} + 0.25 \mathcal{L}_{b}.
            \label{eq:fabulightasd-totalloss}
        \end{equation}

\section{Experiments and Analyses}\label{sec:experiments} 

    In this section, we present a detailed account of the experiments conducted to evaluate the performance of FabuLight-ASD. We first outline the implementation details and then compare FabuLight-ASD to the baseline Light-ASD model using the WASD dataset. Next, we provide a detailed performance breakdown to understand the impact of various factors such as face resolution, pose estimation confidence, and the number of speakers on the model's performance. Finally, we analyse the efficiency of the models in terms of parameter count and multiply-accumulate (MAC) operations, demonstrating that the additional computational cost is negligible.

    \subsection{Implementation Details}\label{sec:experiments-implementation}

        For the following experiments, we utilised the default parameters of Light-ASD. Namely, batches were built by combining videos with the same number of frames as long as the total number of their frames did not surpass 2000, training was performed until a maximum of 30 epochs, FabuLight-ASD was optimised with ADAM with an initial learning rate of $10^{-3}$ and learning decay rate of $0.05$ per epoch.
        
    \subsection{Evaluation of FabuLight-ASD}\label{sec:experiments-fabulight}
        
        The results presented in Table~\ref{tab:experiments-comparisonfabulight} provide a comparative analysis between the performances of two variants of FabuLight-ASD -- one that utilises information from the whole body of a target individual and one that employs information only from their upper body -- and ASD models evaluated by Roxo et al.~\cite{roxo2024} on the WASD dataset. Among these models, Roxo et al.~\cite{roxo2024} include two adaptations of TalkNet, denoted as TalkNet (body + audio) and TalkNet (face + body + audio). The former adaptation receives as input a stack of greyscale body crop images and the audio of the video clip but no face crops, whereas the latter uses input data from all three modalities. It is worth noting that Roxo et al.~\cite{roxo2024} do not provide the overall mean average precision (mAP) performance of those adaptations in their paper.
        
        We evaluated the performance of both FabuLight-ASD variants on the WASD dataset in terms of mAP across different categories. Notably, both variants achieved higher overall performance than every other model, indicating the benefit of incorporating body pose information into the task of determining whether a given person is the source of some perceived speech activity. Furthermore, the upper-body variant presents a slight enhancement in overall performance compared to the whole-body variant. FabuLight-ASD outperforms every other model in each video category, except for surveillance settings (SS), which are the most challenging scenarios in WASD. In this category, TalkNet (face + body + audio) achieves the highest performance with 81.5\% mAP, compared to 77.3\% and 77.1\% for the upper-body and whole-body variants of FabuLight-ASD, respectively. This high performance in surveillance settings, however, occurs at the expense of the model's performance in the remaining categories. Additionally, both TalkNet adaptations are very inefficient compared to FabuLight-ASD, as they include additional cross-attention modules that considerably increase the number of multiply-accumulate (MAC) operations and the models' number of parameters.
        
        Overall, results indicate the benefit of body pose information across all video categories. The noticeable improvements shown by FabuLight-ASD in conditions with speech impairment (SI), face occlusion (FO), and human voice noise (HVN) suggest that the inclusion of body pose information as a spatial-temporal body pose graph is particularly useful in these scenarios. While stacks of greyscale body crop images improve performance in surveillance settings, they do so at the expense of performance in other video categories. Moreover, the use of spatial-temporal body pose graphs not only enhances performance in most scenarios but also keeps the model lightweight, highlighting the advantages of FabuLight-ASD.

        \begin{table*}[!thbp]
            \centering
            \captionsetup{justification=justified}
            \caption{Performance breakdown across various conditions, where ``None'' in the Body column corresponds to Light-ASD, and ``Upper'' and ``Whole'' correspond to the two variants of FabuLight-ASD.}
            \tablebodyfont
            \begin{tabular}{lccccccc}
                \toprule
                Ablation & Body & OC & SI & FO & HVN & SS & Overall \\
                \midrule
                \multicolumn{8}{l}{\textbf{Face resolution}} \\
                \multirow{3}{*}{Small} & None & 96.9 & \textbf{99.9} & \textbf{93.7} & 48.1 & \textbf{77.7} & \textbf{81.7} \\[0.75pt]
                & Upper & \textbf{97.3} & 99.8 & 92.2 & \textbf{60.5} & 75.5 & 80.2 \\[0.75pt]
                & Whole & \textbf{97.3} & \textbf{99.9} & 92.0 & 59.7 & 77.2 & \textbf{81.7} \\
                \cmidrule(lr){2-8}
                \multirow{3}{*}{Medium} & None & 97.9 & 96.7 & 93.4 & 86.8 & 78.4 & 93.0 \\[0.75pt]
                & Upper & \textbf{98.0} & 97.2 & 95.2 & 88.0 & \textbf{79.7} & 93.7 \\[0.75pt]
                & Whole & \textbf{98.0} & \textbf{97.4} & \textbf{95.5} & \textbf{88.6} & 77.6 & \textbf{93.9} \\
                \cmidrule(lr){2-8}
                \multirow{3}{*}{Large} & None & 97.8 & 98.9 & \textbf{97.5} & 84.4 & \textbf{77.3} & 97.2 \\[0.75pt]
                & Upper & \textbf{98.3} & \textbf{99.3} & 97.3 & 84.6 & 73.2 & \textbf{97.4} \\[0.75pt]
                & Whole & 97.6 & 99.2 & 97.3 & \textbf{85.5} & 72.2 & 97.3 \\
                \midrule
                \multicolumn{8}{l}{\textbf{Upper-body pose estimation confidence}} \\
                \multirow{3}{*}{Low} & None & \textbf{88.3} & \textbf{97.5} & \textbf{97.0} & \textbf{79.6} & 64.1 & 82.8 \\[0.75pt]
                & Upper & 87.9 & 96.4 & 96.4 & 76.0 & \textbf{68.8} & \textbf{83.8} \\[0.75pt]
                & Whole & 86.5 & 96.5 & 95.4 & 77.6 & 56.6 & 80.3 \\
                \cmidrule(lr){2-8}
                \multirow{3}{*}{Medium} & None & 95.5 & 98.5 & 96.2 & 86.1 & 72.9 & 93.4 \\[0.75pt]
                & Upper & \textbf{96.8} & \textbf{98.8} & \textbf{96.4} & 86.5 & \textbf{73.2} & 93.8 \\[0.75pt]
                & Whole & 95.2 & \textbf{98.8} & 96.1 & \textbf{87.1} & 72.6 & \textbf{93.9} \\
                \cmidrule(lr){2-8}
                \multirow{3}{*}{High} & None & 98.5 & 97.2 & 94.8 & 83.2 & \textbf{84.5} & 94.3 \\[0.75pt]
                & Upper & \textbf{98.6} & \textbf{97.9} & 95.7 & 85.1 & 82.1 & 94.6 \\[0.75pt]
                & Whole & 98.5 & \textbf{97.9} & \textbf{96.1} & \textbf{86.0} & 83.8 & \textbf{95.2} \\
                \midrule
                \multicolumn{8}{l}{\textbf{Whole-body pose estimation confidence}} \\
                \multirow{3}{*}{Low} & None & 95.5 & 98.7 & 97.6 & 84.4 & 71.4 & 95.6 \\[0.75pt]
                & Upper & \textbf{96.9} & \textbf{98.9} & \textbf{97.9} & \textbf{85.0} & \textbf{74.2} & \textbf{96.1} \\[0.75pt]
                & Whole & 94.7 & \textbf{98.9} & 97.4 & \textbf{85.0} & 71.3 & 95.8 \\
                \cmidrule(lr){2-8}
                \multirow{3}{*}{Medium} & None & 98.1 & 97.3 & 94.6 & 85.3 & \textbf{79.6} & 92.8 \\[0.75pt]
                & Upper & \textbf{98.4} & \textbf{97.9} & 95.4 & 86.2 & 77.4 & 93.0 \\[0.75pt]
                & Whole & 98.2 & \textbf{97.9} & \textbf{95.7} & \textbf{87.5} & 78.8 & \textbf{93.6} \\
                \cmidrule(lr){2-8}
                \multirow{3}{*}{High} & None & \textbf{98.1} & --- & \textbf{95.5} & 80.0 & 77.5 & 92.4 \\[0.75pt]
                & Upper & 97.4 & --- & 95.0 & 81.7 & \textbf{79.7} & 92.4 \\[0.75pt]
                & Whole & 97.7 & --- & 95.4 & \textbf{82.5} & 77.4 & \textbf{92.8} \\
                \midrule
                \multicolumn{8}{l}{\textbf{Number of speakers}} \\
                \multirow{3}{*}{Two} & None & 98.4 & 98.7 & 97.4 & 86.3 & \textbf{83.8} & 95.6 \\[0.75pt]
                & Upper & \textbf{98.6} & 99.0 & \textbf{97.8} & 86.6 & \textbf{83.8} & \textbf{96.1} \\[0.75pt]
                & Whole & 98.5 & \textbf{99.1} & 97.6 & \textbf{87.7} & 83.0 & 96.0 \\
                \cmidrule(lr){2-8}
                \multirow{3}{*}{Three} & None & 92.9 & 98.9 & 93.2 & 77.8 & \textbf{74.2} & 90.3 \\[0.75pt]
                & Upper & \textbf{93.4} & \textbf{99.1} & 93.6 & \textbf{81.3} & 72.4 & 90.1 \\[0.75pt]
                & Whole & 91.4 & \textbf{99.1} & \textbf{94.5} & 81.2 & 73.7 & \textbf{91.1} \\
                \cmidrule(lr){2-8}
                \multirow{3}{*}{At least four} & None & 98.2 & 96.8 & 89.4 & 85.0 & 51.8 & 92.7 \\[0.75pt]
                & Upper & \textbf{98.8} & \textbf{97.6} & 90.9 & 86.1 & 53.3 & 93.7 \\[0.75pt]
                & Whole & 98.2 & 97.5 & \textbf{92.1} & \textbf{87.4} & \textbf{54.4} & \textbf{93.9} \\
                \midrule
                \multicolumn{8}{l}{\textbf{Input temporal span}} \\
                \multirow{3}{*}{Short} & None & 95.6 & \textbf{99.5} & --- & 85.2 & \textbf{81.8} & 88.9 \\[0.75pt]
                & Upper & 97.0 & 99.0 & --- & 83.6 & 77.8 & 88.9 \\[0.75pt]
                & Whole & \textbf{97.8} & 98.8 & --- & \textbf{86.5} & 80.2 & \textbf{90.6} \\
                \cmidrule(lr){2-8}
                \multirow{3}{*}{Medium} & None & \textbf{98.5} & \textbf{97.6} & 95.1 & 82.1 & 72.1 & 94.2 \\[0.75pt]
                & Upper & 98.3 & 97.5 & \textbf{96.0} & \textbf{85.2} & \textbf{82.5} & \textbf{95.1} \\[0.75pt]
                & Whole & 97.5 & 97.2 & 95.3 & 84.1 & 61.7 & 93.6 \\
                \cmidrule(lr){2-8}
                \multirow{3}{*}{Long} & None & 97.7 & 98.3 & 95.5 & 85.1 & \textbf{78.3} & 93.7 \\[0.75pt]
                & Upper & 97.6 & \textbf{98.7} & \textbf{96.1} & \textbf{86.7} & 77.4 & \textbf{94.3} \\[0.75pt]
                & Whole & \textbf{98.1} & \textbf{98.7} & \textbf{96.1} & 85.9 & 78.0 & 94.1 \\
                \bottomrule
            \end{tabular}
            \label{tab:experiments-breakdown}
        \end{table*}

    \subsection{Performance Breakdown}\label{sec:sec:experiments-breakdown}

        To gain deeper insights into FabuLight-ASD's performance compared to Light-ASD\footnote{Despite TalkNet (face + body + audio) achieving superior performance in surveillance settings, Light-ASD achieved the highest overall mAP among the models evaluated by Roxo et al.~\cite{roxo2024}. Light-ASD was deliberately included in this section due to its similar architecture to FabuLight-ASD, facilitating a comparative analysis and providing insights into the impact of body pose information across different subsets of the WASD dataset.}, we divided the validation set of WASD into mutually exclusive groups based on key factors: face resolution, pose estimation confidence, the number of individuals in a scene, and the temporal span of the inputs. By examining the mAP across these divisions, we aim to identify how variations in these conditions impact active speaker detection. This approach allows us to better understand the strengths and limitations of FabuLight-ASD in comparison to Light-ASD, and to evaluate the effectiveness of incorporating body pose information. Table~\ref{tab:experiments-breakdown} summarises the performance of Light-ASD and FabuLight-ASD across various subsets of WASD according to the specific ablation criteria detailed in Sections~\ref{sec:experiments-breakdownfaceresolution} to~\ref{sec:experiments-breakdowntemporalspan}.
        
        It is important to note that the ablation criteria might affect videos from different categories in varied ways. Some categories might contain videos with a higher number of individuals that align more with one criterion than another. This results in a different number of samples being evaluated category-wise, which in turn can lead to the overall mAP metric becoming more biased towards videos from the category more represented by a given ablation criterion. Additionally, some ablation criteria may not apply to certain categories. For example, there are no samples in the subset of ``speech impairment'' videos where an individual received a high average confidence score for the estimation of their whole-body joints (cf. Table~\ref{tab:experiments-breakdown}).
        
        \subsubsection{Face Resolution}\label{sec:experiments-breakdownfaceresolution}

            We divide the data into three groups based on face resolution: large faces (widths greater than 128 pixels), middle faces (widths between 64 and 128 pixels), and small faces (widths smaller than 64 pixels). This division helps to understand the impact of face image size on the model's performance. The category-wise mAP is evaluated for each group, allowing us to see how the model handles different face resolutions under varying conditions.

            The results in Table~\ref{tab:experiments-breakdown} indicate a clear benefit in utilising body pose information in scenarios with human voice background noise, especially when the resolution of the face of the target individual is rather small. Light-ASD presents higher performance for videos with face occlusion and in surveillance settings, yet this is somewhat erratic, as its performance is higher when face resolutions are either small or large. On the other hand, when the face of the target individual presents a medium size, it seems that the pose information has a positive impact, especially that of the upper body. Moreover, in the less challenging scenarios (OC and SI), the upper-body version of FabuLight-ASD slightly outperforms Light-ASD regardless of the face resolution.

            In addition, when comparing the upper-body and whole-body variants of FabuLight-ASD, specific trends emerge regarding face resolution. For small face resolutions, both variants perform similarly across most categories, except in HVN, where the upper-body variant shows superior classification accuracy. However, in SS, the whole-body variant performs better, resulting in its overall superiority in small-resolution samples.
            
            For medium or large face resolutions, the trends shift. The whole-body variant performs better in the HVN category, while the upper-body variant excels in SS. In the OC category, both variants perform similarly when the face resolution is medium, but the upper-body variant gains an advantage when the face resolution is large. Overall, both variants perform comparably for medium and large face resolutions.

        \subsubsection{Pose Estimation Confidence}\label{sec:experiments-breakdownposeconfidence}

            For pose estimation confidence, we evaluate the model's performance using two sets of body joints: the upper-body joints and the whole-body joints. Each set is divided based on the average confidence score of the joints: high confidence (average confidence score of the selected set of body joints greater than 0.75), medium confidence (average confidence score between 0.5 and 0.75), and low confidence (average confidence score lower than 0.5). This evaluation reveals how the reliability of pose information affects detection accuracy in different WASD categories.

            As expected, no significant positive impact has been found by utilising human body pose information with a low average confidence score. However, while this generally leads to a lower performance for both variants of FabuLight-ASD compared to Light-ASD, the upper-body variant performs comparatively better in surveillance settings (cf. Table~\ref{tab:experiments-breakdown}). This result may be influenced by the characteristics of the videos in this category. It is also important to note that the surveillance settings category contains a disproportionately higher number of samples with low-confidence body pose estimations compared to other categories. This higher sample size in surveillance settings likely contributes to the upper-body variant's overall higher mAP. In other words, the better performance of the upper-body variant in this category biases its overall results. Thus, while low-confidence pose information does not generally improve performance, its impact on the upper-body variant seems less detrimental in specific contexts like surveillance settings, where pose data might still provide useful contextual cues despite lower confidence scores.
            
            Conversely, the results in Table~\ref{tab:experiments-breakdown} show that the utilisation of medium-confidence and high-confidence upper-body poses is beneficial in determining whether a target individual is the source of some speech activity. Furthermore, although the utilisation of whole-body pose information is not promising when there is medium average confidence in the body joints of the upper body, the whole-body pose information presents a positive impact in active speaker detection if the upper-body joints of the target individual present a high confidence score on average. Counterintuitively, in surveillance settings, Light-ASD outperforms both variants of FabuLight-ASD when there is an average high confidence in the upper-body joints of the target individual. However, this is not reflected in the overall mAP, which indicates that this high performance of Light-ASD might be due to an exceptional case.

            Finally, no useful analysis can be derived from the performance breakdown based on whole-body joint average confidence scores. Given how this criterion is calculated, several cases categorised as low average confidence have been categorised as medium or high average confidence according to the upper-body pose ablation criterion. Consequently, the results for the low-confidence whole-body pose criterion tend to be close to those of the medium-confidence upper-body pose criterion, and to a lesser extent to those of the high-confidence upper-body pose criterion. Similarly, the results for the medium-confidence whole-body pose criterion closely resemble those of the cases that fit into the high-confidence upper-body pose criterion. Furthermore, the number of dataset samples that fit the high-confidence whole-body pose criterion is rather small (cf. the lack of SI samples that fit this criterion), leading to less informative results.

            When comparing the FabuLight-ASD variants based on pose estimation confidence, a few key patterns emerge. For samples with low upper-body confidence, both variants underperform relative to Light-ASD, though the upper-body variant excels in SS and generally outperforms the whole-body variant across most categories, except in HVN. At medium confidence levels, both variants perform similarly overall. Still, the upper-body variant shows clear advantages in OC, FO, and SS. In contrast, the whole-body variant performs better in HVN when whole-body confidence is low.
            
            The whole-body variant proves more effective at high upper-body confidence, especially in challenging categories like FO, HVN, and SS, consistently outperforming the upper-body variant. While the whole-body variant also performs better in the small subset of samples with high whole-body confidence, particularly in HVN, the upper-body variant remains superior in SS. Overall, the whole-body variant tends to perform better as pose confidence increases, particularly in more challenging scenarios.
            
        \subsubsection{Number of Speakers}\label{sec:experiments-breakdownnumberspeakers}

            We divide the WASD validation set based on the number of individuals present in a scene: two individuals, three individuals, and four or more individuals. This analysis reveals how the complexity of the scene, in terms of the number of speakers, influences the model's ability to detect active speakers accurately. The category-wise mAP is calculated for each group to understand the performance variations across different scenarios.

            Results in Table~\ref{tab:experiments-breakdown} reveal an improvement in active speaker detection when either upper- or whole-body pose is employed in non-surveillance videos. Specifically, there is a slight improvement in less challenging scenarios (OC and SI) and a much higher improvement in scenarios with face occlusion and human voice background noise. Although Light-ASD performs better in surveillance settings, this is particularly true in videos with three speakers, but the performance difference is noticeably small in videos with only two speakers. Finally, in videos with four or more speakers, both variants of FabuLight-ASD outperform Light-ASD in every video category, especially in the most challenging scenarios.

            For scenes with two speakers, both FabuLight-ASD variants perform similarly in OC, SI, and FO categories. However, the whole-body variant outperforms in the HVN category, while the upper-body variant excels in SS. Overall, both variants perform comparably in two-speaker scenes.
            
            In scenes with three speakers, the upper-body variant has an advantage in OC samples, but the whole-body variant surpasses it in both FO and SS, leading to better overall performance. In scenes with four or more speakers, the whole-body variant consistently outperforms the upper-body variant, particularly in more challenging categories like FO, HVN, and SS, though the upper-body variant retains an edge in OC samples.

        \subsubsection{Temporal Span}\label{sec:experiments-breakdowntemporalspan}
            
            To examine the effect of input temporal span on model performance, we categorised the WASD validation set into three groups: short (up to 300 frames), medium (301 to 600 frames), and long (more than 600 frames). The temporal span, represented as $T_{f}$ in Section~\ref{sec:lightasd-feature}, corresponds to the size of the input modalities for both FabuLight-ASD and Light-ASD. By evaluating the category-wise mAP for each group, we can assess how varying input durations affect the performance of both models.

            The performance breakdown in Table~\ref{tab:experiments-breakdown} shows that for short temporal spans, whole-body pose data provides greater advantages than upper-body data, particularly in OC and HVN, where the whole-body version of FabuLight-ASD performs best. For medium temporal spans, the upper-body version of FabuLight-ASD outperforms both the whole-body variant and Light-ASD, especially in challenging categories like FO, HVN, and SS. It achieves the highest performance overall across all temporal span groups. For long temporal spans, both FabuLight-ASD variants either outperform or remain competitive with Light-ASD in every category except SS, where Light-ASD holds a slight advantage. Nevertheless, the upper-body version of FabuLight-ASD achieves the highest overall performance for long temporal spans.
            
            These findings indicate that whole-body pose data is more effective for shorter sequences, while upper-body pose data becomes increasingly valuable in medium and long temporal spans, particularly in challenging categories like FO, HVN, and SS. The upper-body version of FabuLight-ASD performs best in medium temporal spans and maintains strong performance in long spans, especially when compared to Light-ASD.
            
            For short temporal spans, the upper-body variant outperforms the whole-body variant in key categories, including OC, HVN, and SS, resulting in better overall performance. For medium temporal spans, the upper-body variant consistently surpasses the whole-body variant across all categories and overall. In long temporal spans, the upper-body variant continues to excel in HVN samples, while the whole-body variant performs better in OC and SS. Overall, both variants perform similarly for long temporal spans.

    \subsection{Model Efficiency}\label{sec:sec:experiments-efficiency}
    
        Light-ASD stands out as a high-performing ASD model known for being efficient both spatially (retaining a small number of parameters -- 1,021 million) and temporally (executing a low number of operations -- 204 million MACs\footnote{This value was calculated with the `calflops' Python package found at \url{https://pypi.org/project/calflops/}.} per input frame). Table~\ref{tab:experiments-efficiency} demonstrates that the inclusion of a pose feature encoder increases the parameter count to 1,300 million, marking a 27.3\% rise solely attributable to the body feature encoder. Notably, the substitution in the detector component has not altered the parameter count. In other words, the addition of one input modality showed no impact on the parameters of the multimodal fusion mechanism, which comprises a bi-directional GRU and a fully connected layer. The increase in parameters is identical in both variants of FabuLight-ASD due to identical architectures.

        \begin{table}[!ht]
            \centering
            \captionsetup{justification=justified}
            \caption{Impact of inclusion of body pose information as an input modality on model efficiency.}
            \tablebodyfont
            \setlength{\tabcolsep}{3pt}
            \begin{tabular}{lcc}
                \toprule
                Model & Params (M) & MACs (M) \\
                \midrule
                Light-ASD & 1,021 & 204 \\
                \textbf{FabuLight-ASD (upper)} & 1,300 & 207 \\
                \textbf{FabuLight-ASD (whole)} & 1,300 & 209 \\
                \bottomrule
            \end{tabular}
            \label{tab:experiments-efficiency}
        \end{table}
        
        The efficiency of both FabuLight-ASD variants is highlighted not only by the modest rise in parameter count compared to Light-ASD but also by the negligible growth in MAC operations. Table~\ref{tab:experiments-efficiency} indicates that incorporating and processing a sequence of whole-body skeleton graphs results in a 2.4\% increase in MAC operations. Furthermore, this increase diminishes to 1.5\% when using information solely from the upper body. These slight increments in MAC operations primarily stem from the vast majority of operations being performed in the convolutions within the face feature encoder. This underscores the benefits of using pose skeleton graphs as a representation of the body pose modality instead of sequences of whole-body images of the target individual, which would significantly increase the number of MAC operations executed by the model. It is important to note that these efficiency comparisons are based solely on the model architecture as evaluated by Liao et al.~\cite{liao2023}, and do not include additional preprocessing steps such as body pose estimation or human face detection, which could impact overall computational efficiency in real-world applications. Nevertheless, the inherently low number of parameters and computational cost of FabuLight-ASD may offset the effects of these additional steps.

\section{Conclusion}\label{sec:conclusion}

    This study underscores the potential of integrating multiple modalities including body pose information to improve the accuracy and robustness of active speaker detection (ASD) models. We have introduced FabuLight-ASD, a lightweight model that combines facial, audio, and body pose data to achieve improved speaker detection performance. Through extensive experiments on the WASD dataset, we demonstrate the effectiveness of incorporating body pose information, particularly in challenging scenarios like multiple speakers or face occlusion. FabuLight-ASD outperforms other ASD models, especially in categories such as face occlusion and human voice background noise, highlighting the significance of body pose data for handling complex conditions while maintaining computational efficiency.

    Our research paves the way for future investigations in several promising directions. These include exploring real-time developments of FabuLight-ASD in social robots to validate its utility in diverse environments, particularly in multiparty human-robot interaction (MHRI) scenarios where the system must identify active speakers among multiple participants. The utilisation of datasets with egocentric perception videos that include active speaker detection tasks~\cite{donlay2021,grauman2022} would also be beneficial for further testing and enhancing the model's performance. Furthermore, leveraging body pose information to reduce reliance on face information could allow for smaller face dimensions as inputs to the face feature encoder, thereby decreasing the model's MAC operations and improving efficiency. Finally, implementing adaptive weighting techniques to account for the varying contributions of each modality based on their quality could further enhance the model's robustness. These avenues of future work will not only refine FabuLight-ASD but also contribute to the broader field of active speaker detection, ultimately enabling more effective and adaptive human-robot interactions.

\backmatter

\bmhead{Acknowledgements}
The authors thank Cornelius Weber for his valuable feedback on this paper and acknowledge partial support from the German Research Foundation DFG under project CML (TRR 169).

\section*{Declarations}

    \subsection*{Funding}

        This work is partially supported by the German Research Foundation DFG under project CML (TRR 169).

    \subsection*{Competing Interests}

        The authors have no competing interests to declare that are relevant to the content of this article.

    \subsection*{Ethics Approval}

        Not applicable
    
    \subsection*{Consent to Participate}

        Not applicable
    
    \subsection*{Consent for Publication}

        Not applicable
    
    \subsection*{Availability of Data and Materials}

        This work makes use of publicly available data.
    
    \subsection*{Code Availability}

       FabuLight-ASD's code and model weights are available at \url{https://github.com/knowledgetechnologyuhh/FabuLight-ASD}.
    
    \subsection*{Authors' Contributions}

         Conceptualisation: Hugo Carneiro and Stefan Wermter; Methodology: Hugo Carneiro; Formal analysis and investigation: Hugo Carneiro; Writing - original draft preparation: Hugo Carneiro; Writing - review and editing: Stefan Wermter; Funding acquisition: Stefan Wermter; Resources: Stefan Wermter; Supervision: Stefan Wermter.








\bibliography{sn-bibliography}

\end{document}